\documentclass[runningheads]{llncs}
\usepackage{graphicx}
\usepackage{amsmath,amssymb}
\usepackage{booktabs}
\usepackage{multirow}
\usepackage{array}
\usepackage[table]{xcolor}
\usepackage{url}
\usepackage[hidelinks]{hyperref}
\usepackage{caption}
\usepackage{bm}
\usepackage{tikz} 
\usepackage{doi}
\usepackage{fontawesome5}
\usepackage[
  width=122mm,left=12mm,paperwidth=146mm,
  height=193mm,top=12mm,paperheight=217mm
]{geometry}
\newcommand{\method}{\textsc{4D Synchronized Fields}}

\newcommand{\bx}{\mathbf{x}}
\newcommand{\bz}{\mathbf{z}}
\newcommand{\bq}{\mathbf{q}}
\newcommand{\bs}{\mathbf{s}}
\newcommand{\bc}{\mathbf{c}}
\newcommand{\bv}{\mathbf{v}}
\newcommand{\br}{\mathbf{r}}
\newcommand{\be}{\mathbf{e}}

\newcommand{\bw}{\boldsymbol{\omega}}
\newcommand{\bl}{\boldsymbol{\ell}}
\newcommand{\bt}{\mathbf{t}}
\newcommand{\bR}{\mathbf{R}}
\newcommand{\bA}{\mathbf{A}}
\newcommand{\bI}{\mathbf{I}}
\newcommand{\bW}{\mathbf{W}}

\newcommand{\bF}{\mathbf{F}}
\newcommand{\bE}{\mathbf{E}}
\newcommand{\bff}{\mathbf{f}}
\newcommand{\bftilde}{\tilde{\bff}}
\newcommand{\Dtheta}{\mathcal{D}_\theta}
\newcommand{\Mphi}{\mathcal{M}_\phi}
\newcommand{\Eimg}{E_{\mathrm{img}}}
\newcommand{\Etxt}{E_{\mathrm{txt}}}
\DeclareMathOperator{\norm}{norm}
\definecolor{LightBlue}{HTML}{3498DB}
\definecolor{colSync}{HTML}{D1ECF1}
\definecolor{colLang}{HTML}{E0E2B5}
\definecolor{colRecon}{HTML}{FCC495}
\definecolor{colMotion}{HTML}{A58767}
\definecolor{colSyncBest}{HTML}{9DD1E2}
\definecolor{colLangBest}{HTML}{C5CC85}
\definecolor{colReconBest}{HTML}{F0A45C}
\definecolor{colMotionBest}{HTML}{7A5F45}
\newcommand{\rowSync}{\cellcolor{colSync}}
\newcommand{\rowLang}{\cellcolor{colLang}}
\newcommand{\rowRecon}{\cellcolor{colRecon}}

\newcommand{\rowSyncB}{\cellcolor{colSyncBest}}
\newcommand{\rowLangB}{\cellcolor{colLangBest}}
\newcommand{\rowReconB}{\cellcolor{colReconBest}}
\newcommand{\rowMotionB}{\cellcolor{colMotionBest}}
\newcommand{\legendchip}[2]{%
  \tikz[baseline=(X.base)]{%
    \node[fill=#1, rounded corners=2.5pt, inner sep=3pt, minimum height=12pt](X){\strut\scriptsize #2};%
  }%
}
\newcommand{\phasebox}[2]{%
  \par\smallskip\noindent
  \begin{tikzpicture}
    \node[fill=#1, rounded corners=4pt, inner sep=6pt, text width=\dimexpr\linewidth-12pt\relax, align=left]{%
      \small #2%
    };
  \end{tikzpicture}%
  \smallskip\par
}
\newcommand{\secchip}[1]{%
  \tikz[baseline=-0.8ex]{\node[fill=#1, rounded corners=2pt, minimum width=8.5pt, minimum height=8.5pt, inner sep=0pt]{};}%
}
\newcommand{\blueurl}[1]{\href{#1}{\textcolor{LightBlue}{\nolinkurl{#1}}}}
\definecolor{Accent}{RGB}{225,246,237}
\makeatletter
\renewcommand{\@fnsymbol}[1]{\ifcase#1\or\hspace{0pt}\else\hspace{0pt}\fi}
\makeatother

\begin{document}
\pagestyle{headings}
\mainmatter
\title{\method{}: Motion-Language Gaussian Splatting for Temporal Scene Understanding}
\titlerunning{4D Synchronized Fields}
\authorrunning{Barhdadi et al.}
\author{Mohamed Rayan Barhdadi\inst{1} \and Samir Abdaljalil\inst{1} \and Rasul Khanbayov\inst{2} \and \\ Erchin Serpedin\inst{1} \and Hasan Kurban\inst{2}\thanks{Preprint. Contact: \texttt{rayan.barhdadi@tamu.edu} and \texttt{hkurban@hbku.edu.qa}}}
\institute{Texas A\&M University \and Hamad Bin Khalifa University}
\maketitle
\vspace{-12pt}
\begin{center}
{{\footnotesize\faGlobe}\quad\blueurl{Available_soon}}
%{{\footnotesize\faGlobe}\quad\blueurl{https://4d-synchronized-fields.github.io}}
\end{center}
\vspace{-10pt}

% ═════════════════════════════════════════════════════════
%  TEASER
% ═════════════════════════════════════════════════════════
\vspace{-10pt}
\begin{figure}[h]
\centering
\includegraphics[width=0.85\linewidth]{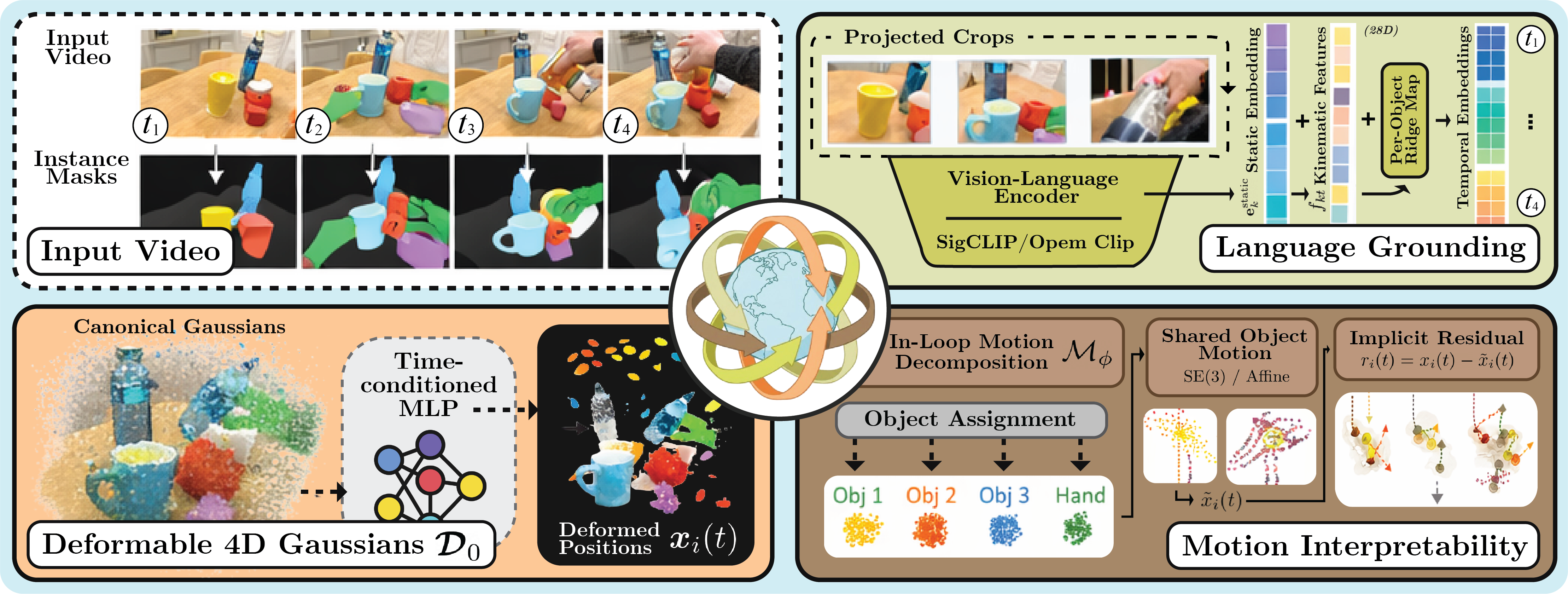}
\captionsetup{margin=1cm}
\caption{\small\textbf{\method{}} learns a deformable 4D Gaussian scene whose per-Gaussian trajectories are decomposed in-loop into shared object motion plus implicit residuals.
The resulting synchronized tracks and kinematics condition an object-time language field, trained from projected object crops and a per-object ridge map, enabling open-vocabulary temporal queries that retrieve both \emph{objects} and \emph{moments}, grounded in learned motion structure.}
\label{fig:teaser}
\end{figure}

% ═════════════════════════════════════════════════════════
%  ABSTRACT
% ═════════════════════════════════════════════════════════
\vspace{-28pt}

\begin{abstract}
Current 4D representations decouple geometry, motion, and semantics: reconstruction methods discard interpretable motion structure; language-grounded methods attach semantics after motion is learned, blind to how objects move; and motion-aware methods encode dynamics as opaque per-point residuals without object-level organization. We propose \textbf{4D Synchronized Fields}, a 4D Gaussian representation that learns object-factored motion in-loop during reconstruction and synchronizes language to the resulting kinematics through a per-object conditioned field. Each Gaussian trajectory is decomposed into shared object motion plus an implicit residual, and a kinematic-conditioned ridge map predicts temporal semantic variation, yielding a single representation in which reconstruction, motion, and semantics are structurally coupled and enabling open-vocabulary temporal queries that retrieve both objects and moments. On HyperNeRF, 4D Synchronized Fields achieves 28.52\,dB mean PSNR, the highest among all language-grounded and motion-aware baselines, within 1.5\,dB of reconstruction-only methods. On targeted temporal-state retrieval, the kinematic-conditioned field attains 0.884 mean accuracy, 0.815 mean vIoU, and 0.733 mean tIoU, surpassing LangSplat (0.415, 0.304, and 0.262) and 4D LangSplat (0.620, 0.433, and 0.439 respectively). Ablation confirms that kinematic conditioning is the primary driver, accounting for $+$0.45 tIoU over a static-embedding-only baseline. 4D Synchronized Fields is the only method that jointly exposes interpretable motion primitives and temporally grounded language fields from a single trained representation. Code will be released.
\keywords{4D Gaussian Splatting, Dynamic Scene Understanding, Object Motion Decomposition, Open-Vocabulary Temporal Queries}
\end{abstract}

% ═════════════════════════════════════════════════════════
%  1. INTRODUCTION
% ═════════════════════════════════════════════════════════
\section{Introduction}
\label{sec:intro}

Building faithful representations of the world is a long-standing goal of visual computing, pursued through stages that successively close the gap between what is represented and what the world actually contains.
Early work focused on what could be extracted from a single image: edges and local features~\cite{marr1982vision,lowe2004sift}, part-based shape models~\cite{brooks1979generalized,felzenszwalb2005pictorial}, and learned hierarchies for recognition at scale~\cite{krizhevsky2012imagenet}, yet recognizing \emph{what} is in an image leaves open the question of \emph{where} things are in three dimensions.
Multi-view geometry and structure-from-motion~\cite{hartley2003multiview,schoenberger2016sfm} recovered 3D structure from image collections; neural radiance fields~\cite{mildenhall2020nerf} encoded scenes as continuous volumetric functions for photorealistic novel-view synthesis~\cite{barron2021mipnerf,muller2022instantngp}; and 3D Gaussian Splatting~\cite{kerbl2023gs} replaced implicit volumetric queries with explicit, rasterizable primitives at real-time rates.
\phasebox{colRecon}{\#1: the world can now be reconstructed and rendered with high fidelity, but the representation is blind to what it depicts.}

Language-grounded methods~\cite{kerr2023lerf,qin2024langsplat,ye2024gaussiangrouping,zhou2024feature3dgs,cen2023saga} attached open-vocabulary semantics to 3D structures.
4D LangSplat~\cite{li2025_4dlangsplat} and related works~\cite{lu2025segment,ji2024sa4d} extended this to temporal representations, but language features are typically distilled \emph{after} motion has been optimized, so the semantic field inherits no structured knowledge of \emph{how} objects move, only a static association with \emph{what} is present.
\phasebox{colLang}{\#2: language grounding lets the representation know \emph{what} it depicts, but it treats motion as a black box and cannot express \emph{how} things move.}

Knowing \emph{what} is not enough when the world is not static: objects move, interact, and change state.
Defor\-mation-based NeRFs~\cite{pumarola2021dnerf,park2021nerfies,park2021hypernerf}, scene-flow methods~\cite{li2021nsff,gao2021dynamicview,liu2023rdrf}, factored space-time representations~\cite{fridovichkeil2023kplanes,cao2023hexplane}, and 4D Gaussian Splatting~\cite{wu2024gs4d,yang2024deformable3dgs,lin2024gaussianflow,li2024stgs} brought dynamic reconstruction within reach, yet motion is encoded as per-point deformation optimized purely for photometric error, exposing no interpretable object-level structure.
\phasebox{colMotion}{\color{white}\#3: motion-aware methods model dynamics, but motion remains an opaque per-point residual with no object-level structure or semantic coupling.}

The absence of object-level structure is not just an engineering limitation; it conflicts with how biological perception is organized.
From the first months of life, infants perceive objects as cohesive bodies by attending to how surfaces move, well before relying on appearance~\cite{spelke1990principles,spelke2000core,carey2009origin}, and objects individuated through motion serve as the hubs around which perception, action, and language coalesce~\cite{smith2005sixlessons,smith2018curriculum}.
World models that must predict, plan, and explain in terms of objects require representations exposing motion as a first-class, interpretable quantity~\cite{sutton2019bitterlesson,sitzmann2026bitterlessoncv,hafner2025dreamerv3,song2025historyguidance}.\footnote{We adopt a perception-first viewpoint: the representation is learned from visual supervision; language serves as an interface to the learned structure over space and time.
This mirrors developmental findings in which infants first build core object representations from spatiotemporal cues and only later integrate linguistic labels~\cite{carey2009origin,waxman1995words}.}
\phasebox{colSync}{\#4: closing the loop requires \emph{synchronization}: motion must be factored by object, and semantics must be conditioned on that factored motion, so that reconstruction, language, and dynamics are unified within a single representation.}

We present \method{}, a 4D scene representation built on the synchronization principle of \#4.
\method{} learns renderable Gaussians while fitting an object-consistent motion model \emph{in-loop} during reconstruction, so that motion structure, object identity, and semantics emerge together rather than being bolted on in sequence.

In summary, our contributions are as follows:
\begin{itemize}
\item[] \textbf{A synchronized 4D scene representation.}
We propose the first representation that synchronizes reconstruction, object-factored motion, and language within a unified Gaussian representation via staged training: motion structure is learned first, and language is then grounded on the resulting kinematics, culminating the progression of \#1 to \#3 and instantiating the synchronization principle of \#4.

\item[] \textbf{In-loop motion decomposition.}
Each Gaussian's trajectory is decomposed into shared object motion (SE(3) or affine) plus an implicit residual, with a residual-adaptive modulation scheme that relaxes regularization on persistently non-rigid Gaussians while leaving the forward renderer unchanged.

\item[] \textbf{Kinematic-conditioned language field.}
We train a per-object ridge map from a kinematic feature vector to semantic residuals, yielding an object-time embedding field that supports open-vocabulary temporal queries and surpasses existing language-grounded baselines on state retrieval.

\item[] \textbf{Structured temporal scene understanding.}
The representation produces synchronized object tracks, motion primitives, interaction graphs, and language slots that a multimodal LLM can directly consume at inference time for temporal reasoning, without having entered the training loop or requiring task-specific retraining.
\end{itemize}

% ═════════════════════════════════════════════════════════
%  2. RELATED WORK
% ═════════════════════════════════════════════════════════
\section{Related Work}
\label{sec:related}
\noindent
We organize prior work along four capability axes and assign each a visual marker used consistently throughout the paper, in section headers, method stages, tables, and result discussions:
\legendchip{colRecon}{Reconstruction}\;\;
\legendchip{colLang}{Language grounding}\;\;
\legendchip{colMotion}{\color{white}Motion structure}\;\;
\legendchip{colSync}{Synchronized (ours)}.
No existing method spans all four; \method{} is the first to do so within a single trained representation.
\\
\noindent
\textbf{3D and 4D reconstruction.}\secchip{colRecon}
NeRFs~\cite{mildenhall2020nerf,barron2021mipnerf,muller2022instantngp} and 3D Gaussian Splatting~\cite{kerbl2023gs} achieve photorealistic novel-view synthesis from static scenes.
Dynamic extensions, including deformation-based NeRFs~\cite{pumarola2021dnerf,park2021nerfies,park2021hypernerf}, scene-flow methods~\cite{li2021nsff,gao2021dynamicview,liu2023rdrf}, factored representations~\cite{fridovichkeil2023kplanes,cao2023hexplane}, and 4D Gaussian variants~\cite{wu2024gs4d,yang2024deformable3dgs,lin2024gaussianflow,li2024stgs}, optimize deformation for photometric fidelity alone, exposing no interpretable object-level kinematics.
\method{} constrains this deformation field with an in-loop object-level factorization (\S\ref{sec:method}).
\\
\noindent
\textbf{Language-grounded scene representations.}\secchip{colLang}
Vision-language features~\cite{radford2021clip,zhai2023siglip,cherti2023openclip} have been distilled into 3D via LERF~\cite{kerr2023lerf}, LangSplat~\cite{qin2024langsplat}, Feature~3DGS~\cite{zhou2024feature3dgs}, and Gaussian Grouping~\cite{ye2024gaussiangrouping}.
In dynamic scenes, 4D LangSplat~\cite{li2025_4dlangsplat} learns dual semantic fields from CLIP and MLLM-generated captions; 4-LEGS~\cite{fiebelman2024legs} and DGD~\cite{labe2024dgd} lift features onto dynamic Gaussians.
In all cases, language is learned on a frozen deformation backbone with no access to object-level motion.
\method{} fits motion primitives during reconstruction and conditions language on the resulting kinematics.
\\
\noindent
\textbf{Motion and object-centric representations.}\secchip{colMotion}
No existing method imposes a training-time factorization of shared object motion versus per-Gaussian residuals; \method{} introduces this \emph{inside} the optimization loop.
Foundation-model segmentation~\cite{kirillov2023sam,ravi2024sam2,carion2025sam3} and dense correspondence~\cite{doersch2022tapvid,doersch2023tapir,karaev2023cotracker} provide post-hoc object signals~\cite{cen2023sam3d,cen2023saga,ji2024sa4d,lu2025segment} but do not shape the motion representation itself.
Motion-graph approaches require skeletal priors; \method{} trades part-level granularity for generality via the modular $\Mphi$ interface (\S\ref{sec:discussion}).
\\
\noindent
\textbf{Toward synchronized temporal understanding.}\secchip{colSync}
Multimodal LLMs~\cite{li2023blip2,liu2023llava}, world models~\cite{hafner2025dreamerv3,song2025historyguidance,chen2025largevideoplanner}, and developmental evidence~\cite{carey2009origin,smith2005sixlessons,smith2018curriculum} all point toward motion-factored object representations as a prerequisite for higher-level reasoning.
Concurrent feed-forward approaches~\cite{wang2025vggsfm} trade per-scene fidelity for scalability; \method{} is, to our knowledge, the first 4D representation that synchronizes reconstruction, motion, and language within a single trained structure.
% ═════════════════════════════════════════════════════════
%  3. METHOD
% ═════════════════════════════════════════════════════════

\section{Method}
\label{sec:method}

We describe \method{} in five stages:
deformable 4D Gaussian reconstruction\enspace\secchip{colRecon} (\S\ref{sec:prelim}),
object assignment\enspace\secchip{colMotion} (\S\ref{sec:assignment}),
in-loop motion decomposition\enspace\secchip{colMotion} (\S\ref{sec:motion_decomp}),
anti-degeneracy objectives with residual-adaptive modulation\enspace\secchip{colMotion} (\S\ref{sec:losses}),
and a synchronized object-time language field\enspace\secchip{colLang} (\S\ref{sec:lang_field}).
Diagnostics and structured temporal export\enspace\secchip{colSync} are described in \S\ref{sec:diagnostics}.
The ordering is deliberate: motion structure must be learned before language can be conditioned on it (\S\ref{sec:lang_field}).
Colored markers indicate which capability from \S\ref{sec:intro} each stage addresses.

\begin{figure}[t]
\centering
\includegraphics[width=0.95\linewidth]{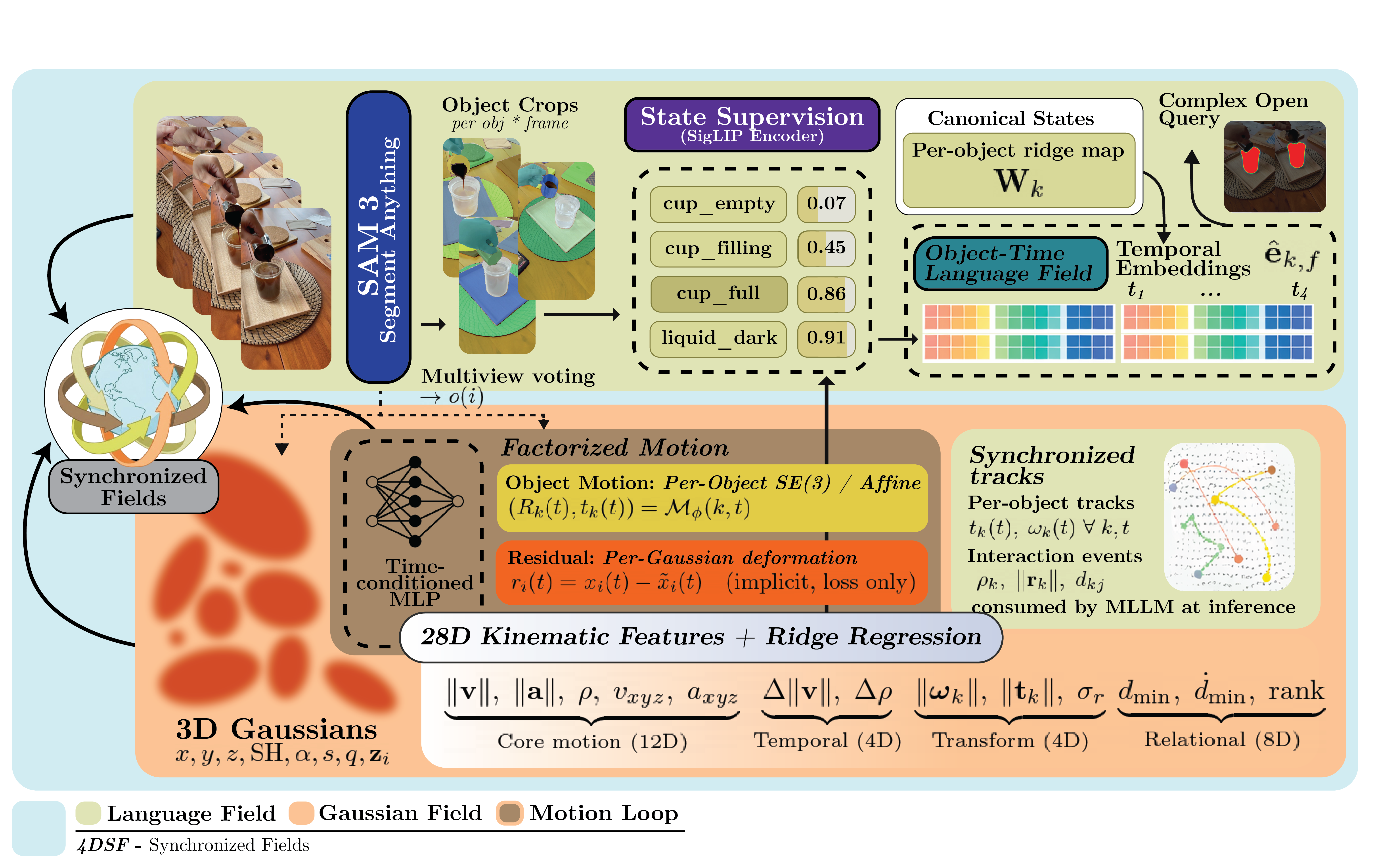}
\caption{\textbf{Method overview.}
A time-conditioned deformation MLP~$\Dtheta$ predicts per-Gaussian deltas yielding deformed positions $\bx_i(t)$.
A shared object-motion model~$\Mphi$ produces per-object transforms yielding object-predicted positions $\tilde{\bx}_i(t)$.
The residual $\br_i(t)=\bx_i(t)-\tilde{\bx}_i(t)$ is defined implicitly and used only in regularizers; rendering uses $\bx_i(t)$ unchanged.
After training, synchronized tracks and 28D kinematic features are extracted and used to fit per-object ridge maps from kinematics to semantic residuals, yielding an object-time language field for open-vocabulary temporal queries; the resulting structured scene description is then leveraged by a multimodal LLM for downstream reasoning.}
\label{fig:pipeline}
\end{figure}

\subsection{Preliminaries: Deformable 4D Gaussian Splatting}
\label{sec:prelim}

\secchip{colRecon}\enspace A scene is represented as $N$ anisotropic Gaussians $\{\mathcal{G}_i\}_{i=1}^N$ with canonical mean $\bx_i^0\!\in\!\mathbb{R}^3$, rotation $\bq_i^0$, log-scale $\bs_i^0$, opacity $\alpha_i$, and SH color $\bc_i$, rendered via alpha-compositing~\cite{kerbl2023gs}.
Each Gaussian carries a learned embedding~$\bz_i\!\in\!\mathbb{R}^D$ that, together with the canonical position and encoded time, conditions a deformation MLP:
\begin{equation}
\label{eq:deform_predict}
(\Delta\bx_i,\;\Delta\bw_i,\;\Delta\bl_i)
\;=\;\Dtheta\!\bigl(\bx_i^0,\,\bz_i,\,\gamma(t)\bigr)\,,
\end{equation}
where $\gamma(t)$ is sinusoidal positional encoding.
The final layer is zero-initialized so that training begins from identity deformation (architecture in Appendix~A).
Deformed parameters at time~$t$ ($\bq(\cdot)$ denotes axis-angle to unit-quaternion conversion):
\begin{align}
\label{eq:deform_pos}
\bx_i(t) &= \bx_i^0 + \Delta\bx_i(t)\,, \\[2pt]
\label{eq:deform_rot}
\bq_i(t) &= \mathrm{normalize}\!\bigl(\bq(\Delta\bw_i(t))\otimes\bq_i^0\bigr)\,, \\[2pt]
\label{eq:deform_scale}
\bs_i(t) &= \exp\!\bigl(\log\bs_i^0 + \Delta\bl_i(t)\bigr)\,.
\end{align}
Training uses the standard photometric loss
\begin{equation} \mathcal{L}_{\mathrm{rgb}}=\lambda_1\|\hat{\bF}-\bF\|_1+\lambda_{\mathrm{dssim}}\,\mathrm{DSSIM}(\hat{\bF},\bF) \end{equation} with standard densification~\cite{kerbl2023gs}.

\subsection{Object Assignment from External Masks}
\label{sec:assignment}

\secchip{colMotion}\enspace The motion decomposition (\S\ref{sec:motion_decomp}) requires each Gaussian to be assigned to an object.
We obtain per-frame instance masks from an external segmenter (SAM~3~\cite{carion2025sam3} in our experiments), though the design is compatible with any instance segmenter.
Each Gaussian~$i$ is projected into every training frame; if it lands on a labeled pixel, one vote is added to that label.
Aggregating votes yields a matrix $\mathbf{V}\!\in\!\mathbb{R}^{N\times(K+1)}$; each Gaussian is assigned to its majority label subject to minimum-vote and minimum-consistency filters (thresholds in Appendix~A).
We write $o(i)\!\in\!\{0,\ldots,K\!-\!1\}$ for the resulting assignment.
Multiview voting suppresses per-frame segmentation noise by requiring consistent labels across many views; we validate assignment quality via a shuffle test in Appendix~C.

\subsection{In-Loop Motion Decomposition}
\label{sec:motion_decomp}

\secchip{colMotion}\enspace Given object assignments from \S\ref{sec:assignment}, \method{} decomposes each Gaussian's predicted motion into (a)~a shared, object-consistent transform and (b)~an implicit per-Gaussian residual, \emph{without altering the forward renderer}.
\\
\noindent
\textbf{Shared object-motion model $\Mphi$.}
A learnable model outputs a rigid or affine transform per object~$k$ and time~$t$.
In the SE(3) parameterization:
\begin{equation}
\label{eq:se3_transform}
\bigl(\bR_k(t),\;\bt_k(t)\bigr)=\Mphi(k,t)\,,
\qquad
\tilde{\bx}_i(t)\;=\;\bR_{o(i)}(t)\,\bx_i^0 + \bt_{o(i)}(t)\,.
\end{equation}
In the affine variant, $\tilde{\bx}_i(t)=(\bI+\Delta\bA_{o(i)}(t))\bx_i^0 + \mathbf{b}_{o(i)}(t)$.
All parameters are initialized near identity (Appendix~A).
\\
\noindent
\textbf{Implicit residual.}
The residual is defined from the predictions of $\Dtheta$ and $\Mphi$:
\begin{equation}
\label{eq:residual}
\br_i(t)\;=\;\bx_i(t)-\tilde{\bx}_i(t)\,.
\end{equation}
By construction $\bx_i(t)=\tilde{\bx}_i(t)+\br_i(t)$, so rendering uses $\bx_i(t)$ unchanged; the decomposition injects training signal only through the regularizers below.

\subsection{Anti-Degeneracy Objectives}
\label{sec:losses}

\secchip{colMotion}\enspace Because the deformation MLP~$\Dtheta$ is strictly more expressive than the per-object SE(3) model, it can absorb all motion into residuals unless explicitly constrained.
We employ five regularizers to prevent this collapse.

\paragraph{(i) Residual energy (base form).}
\begin{equation}
\label{eq:loss_residual}
\mathcal{L}_{\mathrm{res}}
\;=\;\frac{1}{N}\sum_{i=1}^N\|\br_i(t)\|_2^2\,.
\end{equation}

\paragraph{(ii) Residual-adaptive modulation.}
Gaussians on boundaries or articulated joints carry genuinely non-rigid motion and should not be penalized equally.
We track per-Gaussian residual magnitude via EMA ($\eta\!=\!0.95$) and compute a modulation weight from the z-scored running average ($\mathrm{sigm}$: logistic sigmoid):
\begin{equation}
\label{eq:modulation_weight}
m_i = \mathrm{sigm}\!\Bigl(\frac{\bar{r}_i-\mu_{\bar{r}}}{\sigma_{\bar{r}}+\varepsilon}\Bigr)\,,
\qquad
\mathcal{L}_{\mathrm{res}}^{\mathrm{mod}}
\;=\;\frac{1}{N}\sum_{i=1}^N(1-m_i)\,\|\br_i(t)\|_2^2\,.
\end{equation}
$\mathcal{L}_{\mathrm{res}}^{\mathrm{mod}}$ replaces $\mathcal{L}_{\mathrm{res}}$ in the total objective (Eq.~\ref{eq:loss_total}); the base form is shown for clarity.

\paragraph{(iii) Rigid-share hinge.}
We measure the fraction of motion explained by the shared transform and impose a minimum target~$\tau$:
\begin{equation}
\label{eq:rigid_share}
\rho(t)=\frac{1}{N}\sum_{i=1}^N
\frac{\|\tilde{\bx}_i(t)-\bx_i^0\|_2}
     {\|\tilde{\bx}_i(t)-\bx_i^0\|_2+\|\br_i(t)\|_2+\varepsilon}\,,
\qquad
\mathcal{L}_{\mathrm{share}}
=\bigl[\max(0,\;\tau-\rho(t))\bigr]^2\,.
\end{equation}
The hinge is one-sided, active only when $\rho < \tau$, so scenes that naturally exceed $\tau$ incur no penalty.

\paragraph{(iv) Velocity coherence.}
Aligns per-Gaussian and object-level velocity ($t^-$: previous frame):
\begin{equation}
\label{eq:loss_vel}
\mathcal{L}_{\mathrm{vel}}
\;=\;\frac{1}{N}\sum_{i=1}^N
\bigl\|
(\Delta\bx_i(t)-\Delta\bx_i(t^-))
-
(\tilde{\bx}_i(t)-\tilde{\bx}_i(t^-))
\bigr\|_2^2\,.
\end{equation}

\paragraph{(v) Temporal smoothness.}
Penalizes high-frequency variation in the axis-angle $\bw_k(t)$ and translation $\bt_k(t)$ of $\Mphi$ between consecutive frames:
\begin{equation}
\label{eq:loss_smooth}
\mathcal{L}_{\mathrm{smooth}}
\;=\;\frac{1}{K}\sum_{k=0}^{K-1}
\Bigl(\|\bw_k(t)-\bw_k(t^-)\|_2^2 + \|\bt_k(t)-\bt_k(t^-)\|_2^2\Bigr)\,.
\end{equation}

\paragraph{Total objective:}
\begin{equation}
\label{eq:loss_total}
\mathcal{L}
\;=\;\mathcal{L}_{\mathrm{rgb}}
\;+\;\lambda_{\mathrm{res}}\,\mathcal{L}_{\mathrm{res}}^{\mathrm{mod}}
\;+\;\lambda_{\mathrm{share}}\,\mathcal{L}_{\mathrm{share}}
\;+\;\lambda_{\mathrm{vel}}\,\mathcal{L}_{\mathrm{vel}}
\;+\;\lambda_{\mathrm{smooth}}\,\mathcal{L}_{\mathrm{smooth}}\,.
\end{equation}
Motion losses are activated in stages for stability (schedule in Appendix~A).

\subsection{Synchronized Object-Time Language Field}
\label{sec:lang_field}

\secchip{colLang}\enspace The insight driving this stage is that an object's kinematics predict its semantic state: how it moves reveals what it is doing.
We build a semantic embedding field over objects and time from three ingredients: visual crop embeddings, a static appearance anchor, and a per-object ridge map from kinematics to semantic residuals.
The language field is trained on a frozen motion checkpoint rather than jointly with rendering; joint training would couple the high-dimensional SigLIP embedding space to the photometric loss, destabilizing both objectives.
This staged design lets each component converge independently while kinematic conditioning ensures that language is structurally informed by motion.
\\
\noindent
\textbf{Visual observations.}
\label{sec:crop_embed}
For each object~$k$ visible in frame~$f$, we project its Gaussians into the image, extract a bounding-box crop, and encode it:
\begin{equation}
\label{eq:obs_embed}
\be_{k,f}^{\mathrm{obs}}
\;=\;\norm\!\bigl(\Eimg(\mathrm{crop}(k,f))\bigr)\;\in\;\mathbb{R}^{D_e}\,,
\end{equation}
using SigLIP~\cite{zhai2023siglip} by default; we prefer SigLIP over CLIP/OpenCLIP because its sigmoid-based contrastive loss produces better-calibrated per-object similarities for the cross-object scoring in Eq.~\ref{eq:query_score}.
\\
\noindent
\textbf{Static embedding.}
\label{sec:static_embed}
The appearance anchor is the normalized mean over observed frames:
$\be_k^{\mathrm{static}}=\norm\bigl(\frac{1}{|\mathcal{F}_k|}\sum_{f\in\mathcal{F}_k}\be_{k,f}^{\mathrm{obs}}\bigr)$.
\\
\noindent
\textbf{Per-object ridge map.}
\label{sec:ridge}
For each object-time pair $(k,f)$ we construct a 28-dimensional kinematic feature vector $\bff_{k,f}\!\in\!\mathbb{R}^{28}$ encoding core motion, temporal derivatives, transform diagnostics, and relational context (full breakdown in Appendix~B).
We standardize features per object and append a bias term to obtain $\bftilde_{k,f}\!\in\!\mathbb{R}^{29}$, then fit a per-object ridge regression from kinematic features to semantic residuals $\Delta\be_{k,f}\!=\!\be_{k,f}^{\mathrm{obs}}-\be_k^{\mathrm{static}}$:
\begin{equation}
\label{eq:ridge}
\bW_k
\;=\;\arg\min_{\bW}
\sum_{(k,f)\in\mathcal{O}_k}
\bigl\|
\bftilde_{k,f}^{\!\top}\bW - \Delta\be_{k,f}
\bigr\|_2^2
\;+\;\lambda\,\|\bW\|_F^2\,.
\end{equation}
This admits a closed-form solution and is fit independently per object.
\\
\noindent
\textbf{Predicted embedding and blending.}
For all object-time pairs:
\begin{equation}
\label{eq:pred_embed}
\hat{\be}_{k,f}
\;=\;\norm\!\bigl(\be_k^{\mathrm{static}} + \bftilde_{k,f}^{\!\top}\bW_k\bigr)\,.
\end{equation}
Where observations exist, we blend prediction and observation ($\beta\!=\!0.65$); otherwise the prediction is used directly.
A temporal EMA pass ($\gamma\!=\!0.15$) smooths the result; under prolonged occlusion, the kinematic prediction $\hat{\be}_{k,f}$ serves as the sole embedding, gracefully degrading to motion-only semantics.
\\
\noindent
\textbf{Open-vocabulary querying.}
\label{sec:querying}
For query~$q$ with embedding $\be_q\!=\!\norm(\Etxt(q))$, we score each object via a mixture of static and temporal similarity:
\begin{equation}
\label{eq:query_score}
S(k;q)
\;=\;w_s\,\langle\be_k^{\mathrm{static}},\be_q\rangle
\;+\;w_t\,\max_f\,\langle\be_{k,f}^{\mathrm{sync}},\be_q\rangle\,,
\end{equation}
with $w_s\!=\!0.35$ and $w_t\!=\!0.65$.

\subsection{Diagnostics and Structured Export}
\label{sec:diagnostics}

\secchip{colSync}\enspace The decomposition yields four built-in diagnostics (residual mean, rigid-share ratio, assignment shuffle test, and per-object motion variance), detailed in Appendix~C.
\label{sec:reasoning_export}
After training, \method{} exports a structured scene description comprising object slots, per-frame tracks with full kinematics, pairwise interaction events, and synchronized language embeddings (Appendix~D).
This description is consumed at inference time by a multimodal LLM through a dedicated prompt layer (Appendix~G).
% ═════════════════════════════════════════════════════════
%  4. EXPERIMENTS
% ═════════════════════════════════════════════════════════
\section{Experiments}
\label{sec:experiments}
We evaluate \method{} along three axes: reconstruction fidelity, motion factorization quality, and language synchronization.

\subsection{Experimental Setup}
\label{sec:setup}
\noindent
\textbf{Datasets.}
We evaluate on \textbf{HyperNeRF}~\cite{park2021hypernerf} (6~scenes) and \textbf{Neu3D}~\cite{li2022neu3d}.
\\
\noindent
\textbf{Metrics.}
Reconstruction: PSNR, SSIM, LPIPS~\cite{zhang2018lpips}.
Motion: rigid-share ratio~$\rho$, residual mean.
Language: query-state accuracy (Acc), volumetric IoU (vIoU), temporal interval IoU (tIoU).
\\
\noindent
\textbf{Baselines.}
\emph{Language-grounded}: 4D LangSplat~\cite{li2025_4dlangsplat}, LangSplat~\cite{qin2024langsplat}, LEGaussians~\cite{shi2024legaussians}, LERF~\cite{kerr2023lerf};
\emph{reconstruction-only}: Deformable 3DGS~\cite{yang2024deformable3dgs}, SC-GS~\cite{huang2024scgs}, 4DGaussians~\cite{wu2024gs4d}, 3DGS~\cite{kerbl2023gs};
\emph{motion-aware}: MotionGS~\cite{guo2024motiongs}.
All methods are evaluated on a held-out test split at matched resolution under a controlled protocol (details in Table~\ref{tab:recon_fair_x2_sweep} caption and Appendix~A).
4-LEGS~\cite{fiebelman2024legs} and DGD~\cite{labe2024dgd} are discussed in \S\ref{sec:related} but excluded from Tables~\ref{tab:recon_fair_x2_sweep} and \ref{tab:main_lang_comparison} because neither provides public code or pre-trained models on HyperNeRF at the time of submission. Motion-Blender GS targets skeletal scenes and does not report on this benchmark.
\\
\noindent
\textbf{Architecture and runtime.}
$\Dtheta$ is a 4-layer MLP (128-dim hidden, 0.13M params); $\Mphi$ stores per-object per-frame SE(3) parameters (6$K$$T$ floats, $<$0.01M).
Total training: ${\sim}$41\,min per scene on a single A100 (30k backbone iterations + 20k motion decomposition); the language field (closed-form ridge) adds $<$1\,min.
The motion decomposition adds ${\sim}$12\% wall-time overhead relative to the backbone alone, with negligible memory cost since $\Mphi$ operates on object centroids rather than all $N$ Gaussians.

\subsection{Reconstruction Quality}
\label{sec:recon}

Table~\ref{tab:recon_fair_x2_sweep} presents a controlled comparison.
Among language-grounded methods, \method{} achieves the highest mean PSNR (28.52\,dB vs.\ 25.58\,dB for 4D LangSplat) and leads on all four metrics.
Reconstruction-only methods score higher because they optimize exclusively for photometric fidelity with no structural constraints.
Despite five additional regularizers (\S\ref{sec:losses}), the gap to the recon-only upper bound (Deformable-3DGS, 30.04\,dB) is only 1.5\,dB, substantially narrower than the 4.5\,dB gap for 4D LangSplat, indicating that the factorization acts as a beneficial inductive bias rather than a reconstruction penalty.
All methods use identical held-out frames, resolution, and COLMAP initialization; our method uses SigLIP (no MLLM captions during training), while 4D LangSplat additionally uses MLLM-generated captions as supervision.
Runtime and parameter counts are reported in Appendix~A.

% ═════════ FULL RECONSTRUCTION TABLE (UNCHANGED) ═════════
\begin{table*}[ht!]
\centering
\caption{\textbf{Novel-view reconstruction on HyperNeRF} (held-out 4D LangSplat test split, $2\times$ resolution).
Best per category in \emph{darker} shade; our results \textbf{bold}.
Categories are \colorbox{colSync}{\textbf{not directly comparable}}: reconstruction-only methods (orange) optimize purely for pixel fidelity with no motion or language objectives.
Despite additional structural constraints, our method (blue) closes to within 1.5\,dB of the recon-only upper bound while \colorbox{colLang}{\textbf{outperforming all language-grounded and motion-aware baselines}}.}
\label{tab:recon_fair_x2_sweep}
\vspace{2pt}

\footnotesize
\legendchip{colSync}{Ours (synchronized)}\;\;
\legendchip{colLang}{Language-grounded}\;\;
\legendchip{colRecon}{Reconstruction-only}\;\;
\legendchip{colMotion}{\color{white}Motion-aware}
\par\vspace{4pt}

\resizebox{\textwidth}{!}{%
\scriptsize
\renewcommand{\arraystretch}{1.20}
\setlength{\tabcolsep}{3.5pt}
\begin{tabular}{l l ccccccc}
\toprule
Metric & Method & americano & chickchicken & espresso & keyboard & split-cookie & torchocolate & Mean \\
\midrule
\multirow{10}{*}{\rotatebox[origin=c]{90}{PSNR\,$\uparrow$}}
& \rowSyncB \textbf{Ours}    & \rowSyncB \textbf{29.23} & \rowSyncB \textbf{27.49} & \rowSyncB \textbf{27.44} & \rowSyncB \textbf{29.67} & \rowSyncB \textbf{32.70} & \rowSyncB \textbf{24.59} & \rowSyncB \textbf{28.52} \\
& \rowLang 4D LangSplat      & \rowLang 28.00 & \rowLang 26.46 & \rowLang 19.15 & \rowLang 27.13 & \rowLang 27.70 & \rowLang 25.02 & \rowLang 25.58 \\
& \rowLang LangSplat         & \rowLang 16.57 & \rowLang 17.24 & \rowLang 16.71 & \rowLang 16.43 & \rowLang 16.65 & \rowLang 16.57 & \rowLang 16.69 \\
& \rowLang LeGaussians       & \rowLang 16.48 & \rowLang 17.18 & \rowLang 16.55 & \rowLang 16.48 & \rowLang 16.77 & \rowLang 16.57 & \rowLang 16.67 \\
& \rowLangB LERF             & \rowLangB 21.13 & \rowLangB 19.97 & \rowLangB \textbf{24.39} & \rowLangB \textbf{24.44} & \rowLangB 20.82 & \rowLangB 20.79 & \rowLangB 21.92 \\
& \rowReconB Deformable-3DGS & \rowReconB \textbf{32.12} & \rowReconB \textbf{28.02} & \rowReconB \textbf{27.11} & \rowReconB 29.70 & \rowReconB \textbf{34.13} & \rowReconB \textbf{29.18} & \rowReconB \textbf{30.04} \\
& \rowRecon SC-GS            & \rowRecon 31.39 & \rowRecon 27.66 & \rowRecon 26.85 & \rowRecon \textbf{29.95} & \rowRecon 32.97 & \rowRecon 28.69 & \rowRecon 29.59 \\
& \rowRecon 4DGaussians      & \rowRecon 28.92 & \rowRecon 27.07 & \rowRecon 25.44 & \rowRecon 28.26 & \rowRecon 31.10 & \rowRecon 26.32 & \rowRecon 27.85 \\
& \rowRecon 3DGS             & \rowRecon 17.33 & \rowRecon 17.60 & \rowRecon 16.97 & \rowRecon 16.66 & \rowRecon 17.36 & \rowRecon 16.78 & \rowRecon 17.12 \\
& \rowMotionB \color{white}MotionGS & \rowMotionB \color{white}29.62 & \rowMotionB \color{white}27.26 & \rowMotionB \color{white}26.26 & \rowMotionB \color{white}28.43 & \rowMotionB \color{white}30.03 & \rowMotionB \color{white}27.20 & \rowMotionB \color{white}28.13 \\
\midrule
\multirow{10}{*}{\rotatebox[origin=c]{90}{SSIM\,$\uparrow$}}
& \rowSyncB \textbf{Ours}    & \rowSyncB \textbf{0.91} & \rowSyncB \textbf{0.80} & \rowSyncB \textbf{0.92} & \rowSyncB \textbf{0.92} & \rowSyncB \textbf{0.94} & \rowSyncB \textbf{0.87} & \rowSyncB \textbf{0.89} \\
& \rowLangB 4D LangSplat     & \rowLangB 0.87 & \rowLangB 0.78 & \rowLangB 0.85 & \rowLangB \textbf{0.87} & \rowLangB 0.81 & \rowLangB \textbf{0.84} & \rowLangB \textbf{0.84} \\
& \rowLang LangSplat         & \rowLang 0.63 & \rowLang 0.73 & \rowLang 0.80 & \rowLang 0.72 & \rowLang 0.68 & \rowLang 0.76 & \rowLang 0.72 \\
& \rowLang LeGaussians       & \rowLang 0.63 & \rowLang 0.73 & \rowLang 0.80 & \rowLang 0.72 & \rowLang 0.68 & \rowLang 0.75 & \rowLang 0.72 \\
& \rowLang LERF              & \rowLang 0.67 & \rowLang 0.74 & \rowLang \textbf{0.88} & \rowLang 0.83 & \rowLang 0.60 & \rowLang 0.75 & \rowLang 0.75 \\
& \rowReconB Deformable-3DGS & \rowReconB \textbf{0.94} & \rowReconB \textbf{0.84} & \rowReconB \textbf{0.92} & \rowReconB \textbf{0.92} & \rowReconB \textbf{0.95} & \rowReconB \textbf{0.91} & \rowReconB \textbf{0.91} \\
& \rowRecon SC-GS            & \rowRecon 0.93 & \rowRecon 0.84 & \rowRecon 0.92 & \rowRecon 0.92 & \rowRecon 0.94 & \rowRecon 0.90 & \rowRecon 0.91 \\
& \rowRecon 4DGaussians      & \rowRecon 0.86 & \rowRecon 0.82 & \rowRecon 0.90 & \rowRecon 0.89 & \rowRecon 0.88 & \rowRecon 0.86 & \rowRecon 0.87 \\
& \rowRecon 3DGS             & \rowRecon 0.67 & \rowRecon 0.74 & \rowRecon 0.81 & \rowRecon 0.74 & \rowRecon 0.71 & \rowRecon 0.76 & \rowRecon 0.74 \\
& \rowMotionB \color{white}MotionGS & \rowMotionB \color{white}0.91 & \rowMotionB \color{white}0.83 & \rowMotionB \color{white}0.91 & \rowMotionB \color{white}0.90 & \rowMotionB \color{white}0.91 & \rowMotionB \color{white}0.89 & \rowMotionB \color{white}0.89 \\
\midrule
\multirow{10}{*}{\rotatebox[origin=c]{90}{LPIPS-A\,$\downarrow$}}
& \rowSyncB \textbf{Ours}    & \rowSyncB \textbf{0.097} & \rowSyncB \textbf{0.291} & \rowSyncB \textbf{0.145} & \rowSyncB \textbf{0.101} & \rowSyncB \textbf{0.080} & \rowSyncB \textbf{0.205} & \rowSyncB \textbf{0.153} \\
& \rowLang 4D LangSplat      & \rowLang 0.163 & \rowLang 0.330 & \rowLang 0.203 & \rowLang 0.165 & \rowLang 0.294 & \rowLang 0.267 & \rowLang 0.237 \\
& \rowLang LangSplat         & \rowLang 0.589 & \rowLang 0.606 & \rowLang 0.568 & \rowLang 0.627 & \rowLang 0.579 & \rowLang 0.614 & \rowLang 0.597 \\
& \rowLang LeGaussians       & \rowLang 0.588 & \rowLang 0.606 & \rowLang 0.564 & \rowLang 0.627 & \rowLang 0.580 & \rowLang 0.614 & \rowLang 0.596 \\
& \rowLangB LERF             & \rowLangB \textbf{0.141} & \rowLangB \textbf{0.293} & \rowLangB \textbf{0.156} & \rowLangB \textbf{0.165} & \rowLangB \textbf{0.213} & \rowLangB \textbf{0.290} & \rowLangB \textbf{0.210} \\
& \rowReconB Deformable-3DGS & \rowReconB \textbf{0.062} & \rowReconB \textbf{0.224} & \rowReconB \textbf{0.145} & \rowReconB \textbf{0.087} & \rowReconB \textbf{0.056} & \rowReconB \textbf{0.150} & \rowReconB \textbf{0.121} \\
& \rowRecon SC-GS            & \rowRecon 0.069 & \rowRecon 0.253 & \rowRecon 0.152 & \rowRecon 0.090 & \rowRecon 0.066 & \rowRecon 0.184 & \rowRecon 0.136 \\
& \rowRecon 4DGaussians      & \rowRecon 0.153 & \rowRecon 0.295 & \rowRecon 0.181 & \rowRecon 0.188 & \rowRecon 0.184 & \rowRecon 0.300 & \rowRecon 0.217 \\
& \rowRecon 3DGS             & \rowRecon 0.685 & \rowRecon 0.666 & \rowRecon 0.596 & \rowRecon 0.671 & \rowRecon 0.669 & \rowRecon 0.654 & \rowRecon 0.657 \\
& \rowMotionB \color{white}MotionGS & \rowMotionB \color{white}0.100 & \rowMotionB \color{white}0.290 & \rowMotionB \color{white}0.183 & \rowMotionB \color{white}0.118 & \rowMotionB \color{white}0.121 & \rowMotionB \color{white}0.227 & \rowMotionB \color{white}0.173 \\
\midrule
\multirow{10}{*}{\rotatebox[origin=c]{90}{LPIPS-V\,$\downarrow$}}
& \rowSyncB \textbf{Ours}    & \rowSyncB \textbf{0.161} & \rowSyncB \textbf{0.372} & \rowSyncB \textbf{0.249} & \rowSyncB \textbf{0.205} & \rowSyncB \textbf{0.144} & \rowSyncB \textbf{0.284} & \rowSyncB \textbf{0.236} \\
& \rowLang 4D LangSplat      & \rowLang 0.253 & \rowLang 0.433 & \rowLang 0.323 & \rowLang 0.292 & \rowLang 0.373 & \rowLang 0.354 & \rowLang 0.338 \\
& \rowLang LangSplat         & \rowLang 0.586 & \rowLang 0.613 & \rowLang 0.561 & \rowLang 0.610 & \rowLang 0.597 & \rowLang 0.574 & \rowLang 0.590 \\
& \rowLang LeGaussians       & \rowLang 0.587 & \rowLang 0.615 & \rowLang 0.560 & \rowLang 0.609 & \rowLang 0.595 & \rowLang 0.575 & \rowLang 0.590 \\
& \rowLangB LERF             & \rowLangB \textbf{0.141} & \rowLangB \textbf{0.293} & \rowLangB \textbf{0.156} & \rowLangB \textbf{0.165} & \rowLangB \textbf{0.213} & \rowLangB \textbf{0.290} & \rowLangB \textbf{0.210} \\
& \rowReconB Deformable-3DGS & \rowReconB \textbf{0.113} & \rowReconB 0.320 & \rowReconB 0.249 & \rowReconB 0.180 & \rowReconB \textbf{0.104} & \rowReconB \textbf{0.225} & \rowReconB \textbf{0.198} \\
& \rowRecon SC-GS            & \rowRecon 0.121 & \rowRecon 0.353 & \rowRecon 0.264 & \rowRecon 0.186 & \rowRecon 0.116 & \rowRecon 0.258 & \rowRecon 0.216 \\
& \rowRecon 4DGaussians      & \rowRecon 0.249 & \rowRecon 0.393 & \rowRecon 0.291 & \rowRecon 0.279 & \rowRecon 0.257 & \rowRecon 0.362 & \rowRecon 0.305 \\
& \rowRecon 3DGS             & \rowRecon 0.618 & \rowRecon 0.638 & \rowRecon 0.569 & \rowRecon 0.630 & \rowRecon 0.625 & \rowRecon 0.592 & \rowRecon 0.612 \\
& \rowMotionB \color{white}MotionGS & \rowMotionB \color{white}0.158 & \rowMotionB \color{white}0.381 & \rowMotionB \color{white}0.280 & \rowMotionB \color{white}0.210 & \rowMotionB \color{white}0.182 & \rowMotionB \color{white}0.293 & \rowMotionB \color{white}0.251 \\
\bottomrule
\end{tabular}
}% end resizebox
\end{table*}

\subsection{Motion Factorization Quality}
\label{sec:factorization_results}

The factorization achieves rigid-share ratios of $0.41$ to $0.53$ across scenes (mean $\bar{\rho}\!=\!0.472$, mean residual $0.037$).
Scenes with predominantly rigid motion (split-cookie: $0.528$; americano: $0.508$) achieve higher rigid share; scenes with significant non-rigid deformation (espresso: $0.411$; keyboard: $0.422$) have lower rigid share and higher residual, consistent with physical content.
No existing dynamic benchmark provides ground-truth per-object SE(3) trajectories, so motion quality is assessed via internal diagnostics.
The shuffle test (Appendix~C) confirms object-specificity: permuting labels increases residual energy by $1.52\times$.
Across scenes, 23\% of Gaussians have $m_i\!>\!0.5$, concentrated on boundaries and articulated regions as expected.

\subsection{Language Synchronization}
\label{sec:lang_results}

\begin{figure}[t]
\centering
\includegraphics[width=\textwidth]{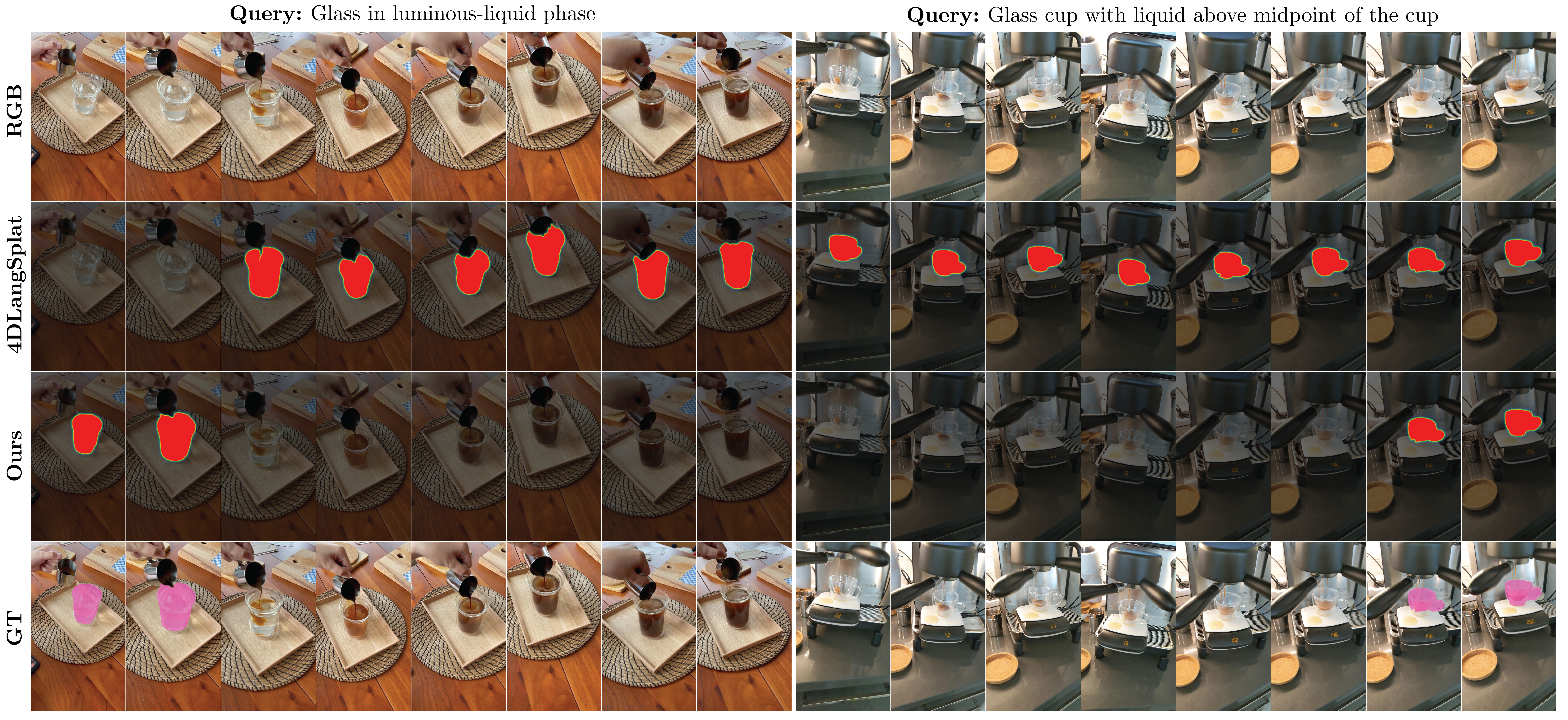}
\caption{\textbf{Targeted temporal-state retrieval} on americano (left: ``glass in luminous-liquid phase'') and espresso (right: ``glass cup with liquid above midpoint'').
Rows: RGB input, 4D LangSplat activation, \method{} activation, ground truth.
4D LangSplat activates broadly across the temporal extent; \method{} produces tighter activations aligned with ground-truth intervals, reflecting kinematic conditioning's sensitivity to motion-correlated state transitions.}
\label{fig:lang_qual}
\end{figure}

Table~\ref{tab:main_lang_comparison} reports targeted temporal-state retrieval on four HyperNeRF scenes with human-annotated per-frame state labels and explicit query-to-state mapping heuristics designed to isolate temporal reasoning precision (e.g., americano: ``luminous-liquid phase''; espresso: ``liquid above midpoint'').
For each state query the method returns a binary per-frame activation; Acc, vIoU, and tIoU are computed against ground-truth frame intervals (Appendix~F).
\method{} achieves 0.884 mean Acc (vs.\ 0.415 for LangSplat, 0.620 for 4D LangSplat), 0.815 mean vIoU (vs.\ 0.304 and 0.433), and 0.733 mean tIoU (vs.\ 0.262 and 0.439).
The largest gains are on americano ($\Delta$Acc$\,{=}\,{+}0.62$, $\Delta$vIoU$\,{=}\,{+}0.86$) and espresso ($\Delta$Acc$\,{=}\,{+}0.30$, $\Delta$vIoU$\,{=}\,{+}0.43$), where state transitions are tightly coupled to object motion; the gap is narrower on chickchicken, where state changes are partly appearance-driven.
Figure~\ref{fig:lang_qual} shows qualitative temporal localization on two challenging queries.

% ═════════ LANGUAGE TABLE (UPDATED: +LangSplat row, new numbers) ═════════
\begin{table*}[ht!]
\centering
\caption{\textbf{Targeted temporal-state retrieval} on four HyperNeRF scenes with human-annotated state labels.
Best per metric in \emph{darker} shade; our results \textbf{bold}.}
\label{tab:main_lang_comparison}
\vspace{2pt}

\footnotesize
\legendchip{colSyncBest}{Ours (synchronized)}\;\;
\legendchip{colLangBest}{Language-grounded baselines}
\par\vspace{4pt}

\resizebox{\textwidth}{!}{%
\scriptsize
\renewcommand{\arraystretch}{1.22}
\setlength{\tabcolsep}{4pt}
\begin{tabular}{l l ccccc}
\toprule
Metric & Method & americano & chickchicken & espresso & split-cookie & Mean \\
\midrule
\multirow{3}{*}{\rotatebox[origin=c]{90}{Acc\,$\uparrow$}}
& \rowSyncB \textbf{Ours} & \rowSyncB \textbf{1.000} & \rowSyncB \textbf{0.717} & \rowSyncB \textbf{0.951} & \rowSyncB \textbf{0.868} & \rowSyncB \textbf{0.884} \\
& \rowLang LangSplat      & \rowLang 0.885            & \rowLang 0.500            & \rowLang 0.086            & \rowLang 0.189            & \rowLang 0.415 \\
& \rowLangB 4D LangSplat  & \rowLangB 0.385           & \rowLangB 0.630           & \rowLangB 0.654           & \rowLangB 0.811           & \rowLangB 0.620 \\
\midrule
\multirow{3}{*}{\rotatebox[origin=c]{90}{vIoU\,$\uparrow$}}
& \rowSyncB \textbf{Ours} & \rowSyncB \textbf{1.000} & \rowSyncB \textbf{0.683} & \rowSyncB \textbf{0.810} & \rowSyncB \textbf{0.767} & \rowSyncB \textbf{0.815} \\
& \rowLang LangSplat      & \rowLang 0.700            & \rowLang 0.395            & \rowLang 0.000            & \rowLang 0.122            & \rowLang 0.304 \\
& \rowLangB 4D LangSplat  & \rowLangB 0.135           & \rowLangB 0.541           & \rowLangB 0.378           & \rowLangB 0.677           & \rowLangB 0.433 \\
\midrule
\multirow{3}{*}{\rotatebox[origin=c]{90}{tIoU\,$\uparrow$}}
& \rowSyncB \textbf{Ours} & \rowSyncB \textbf{0.929} & \rowSyncB \textbf{0.418} & \rowSyncB \textbf{0.821} & \rowSyncB \textbf{0.762} & \rowSyncB \textbf{0.733} \\
& \rowLang LangSplat      & \rowLang 0.777            & \rowLang 0.220            & \rowLang 0.000            & \rowLang 0.052            & \rowLang 0.262 \\
& \rowLangB 4D LangSplat  & \rowLangB 0.149           & \rowLangB 0.290           & \rowLangB 0.592           & \rowLangB 0.724           & \rowLangB 0.439 \\
\bottomrule
\end{tabular}
}% end resizebox
\end{table*}

\subsection{Ablation Studies}
\label{sec:ablation}
\noindent
\textbf{Reconstruction backbone.}
Tables~\ref{tab:abl_components} and \ref{tab:abl_pipeline} present component and pipeline ablations.
Removing opacity reset costs $-$0.08\,dB; further delaying densification costs $-$0.24\,dB total.
The full pipeline recovers +0.87\,dB over the backbone alone (Table~\ref{tab:abl_pipeline}).
\\
\noindent
\textbf{Motion decomposition components.}
Removing $\Mphi$ eliminates the factorization ($\bar{\rho}\!\to\!0$) and increases residual energy by 32\%, yet PSNR is unchanged ($+$0.07\,dB), confirming interpretable structure at zero reconstruction cost.
Removing modulation degrades PSNR ($-$0.11\,dB) and inflates residuals by 52\%.
Removing the hinge collapses $\bar{\rho}$ by 37\%.
The staged loss activation is essential: simultaneous activation causes $\bar{\rho}$ to plateau below 0.15.
The target $\tau\!=\!0.35$ is fixed across all scenes; the one-sided hinge means scenes exceeding it (split-cookie: $\rho\!=\!0.53$) incur zero penalty.
\\
\noindent
\textbf{Kinematic conditioning.}
Table~\ref{tab:abl_kinematic} isolates the effect of kinematic conditioning by comparing the full ridge-map pipeline against a static-embedding-only baseline.
Removing kinematic conditioning drops mean Acc from 0.884 to 0.626 ($-$0.258), vIoU from 0.815 to 0.446 ($-$0.369), and tIoU from 0.733 to 0.279 ($-$0.453), more than halving temporal localization quality.
The largest drops are on espresso ($\Delta$Acc$\,{=}\,{-}0.74$, $\Delta$tIoU$\,{=}\,{-}0.60$) and americano ($\Delta$vIoU$\,{=}\,{-}0.64$, $\Delta$tIoU$\,{=}\,{-}0.64$), both motion-dominated.
Even on chickchicken, kinematic conditioning improves tIoU by $+$0.31.
This confirms that kinematic features are the primary driver of temporal localization: the static anchor alone cannot distinguish \emph{when} a state holds.

% ── Side-by-side component + pipeline ablation tables (UNCHANGED) ──
\begin{table}[t]
\begin{minipage}[t]{0.50\linewidth}
\centering
\caption{\small Component ablation (4-scene mean). $\bar{\rho}$: rigid-share ratio; Res.: residual L2.}
\label{tab:abl_components}
\vspace{2pt}
\scriptsize
\setlength{\tabcolsep}{2pt}
\renewcommand{\arraystretch}{1.18}
\begin{tabular}{lcccc}
\toprule
Variant & $\Delta$PSNR & $\Delta$SSIM & $\bar{\rho}$ & Res. \\
\midrule
\rowSyncB \textbf{Ours} & \rowSyncB \textbf{28.52} & \rowSyncB \textbf{.89} & \rowSyncB \textbf{.472} & \rowSyncB \textbf{.037} \\
\midrule
\multicolumn{5}{l}{\cellcolor{colSync}\emph{Backbone (no motion decomp.)}} \\
\rowSync w/o op.\ reset     & \rowSync $-$0.08 & \rowSync $-$.000 & \rowSync & \rowSync \\
\rowSync Low reset           & \rowSync $-$0.05 & \rowSync +.000  & \rowSync & \rowSync \\
\rowSync w/o reset+late den. & \rowSync $-$0.24 & \rowSync $-$.002 & \rowSync & \rowSync \\
\midrule
\multicolumn{5}{l}{\cellcolor{colSync}\emph{Motion decomposition}} \\
\rowSync w/o $\Mphi$         & \rowSync +0.07 & \rowSync +.000 & \rowSync .000 & \rowSync +.002 \\
\rowSync w/o modulation      & \rowSync $-$0.11 & \rowSync $-$.002 & \rowSync .330 & \rowSync +.004 \\
\rowSync w/o share hinge     & \rowSync +0.03 & \rowSync $-$.007 & \rowSync .209 & \rowSync $-$.003 \\
\bottomrule
\end{tabular}
\end{minipage}
\hfill
\begin{minipage}[t]{0.46\linewidth}
\centering
\caption{\small Pipeline stages (4-scene mean). \checkmark\,=\,capability.}
\label{tab:abl_pipeline}
\vspace{2pt}
\scriptsize
\setlength{\tabcolsep}{2pt}
\renewcommand{\arraystretch}{1.18}
\begin{tabular}{lcccc}
\toprule
Stage & PSNR & Mot. & Lang. & Exp. \\
\midrule
\rowSyncB \textbf{Ours} & \rowSyncB \textbf{28.52} & \rowSyncB \checkmark & \rowSyncB \checkmark & \rowSyncB \checkmark \\
\midrule
\rowRecon Backbone only & \rowRecon 27.59 & \rowRecon $\times$ & \rowRecon $\times$ & \rowRecon $\times$ \\
\rowMotionB \color{white}+ motion & \rowMotionB \color{white}28.46 & \rowMotionB \color{white}\checkmark & \rowMotionB \color{white}$\times$ & \rowMotionB \color{white}$\times$ \\
\rowSync + language & \rowSync 28.46 & \rowSync \checkmark & \rowSync \checkmark & \rowSync \checkmark \\
\bottomrule
\end{tabular}
\end{minipage}
\end{table}

% ── NEW: Kinematic conditioning ablation table ──
\begin{table}[t]
\centering
\caption{\small Kinematic conditioning ablation (4-scene mean). Removing the ridge map from kinematics to semantics.}
\label{tab:abl_kinematic}
\vspace{2pt}
\scriptsize
\setlength{\tabcolsep}{3pt}
\renewcommand{\arraystretch}{1.18}
\begin{tabular}{lccc}
\toprule
Variant & Acc & vIoU & tIoU \\
\midrule
\rowSyncB \textbf{Full (ridge + kinematics)} & \rowSyncB \textbf{0.884} & \rowSyncB \textbf{0.815} & \rowSyncB \textbf{0.733} \\
\rowSync Static embedding only & \rowSync 0.626 & \rowSync 0.446 & \rowSync 0.279 \\
\midrule
\rowSync $\Delta$ (drop) & \rowSync $-$0.258 & \rowSync $-$0.369 & \rowSync $-$0.453 \\
\bottomrule
\end{tabular}
\end{table}

\section{Discussion}
\label{sec:discussion}
\noindent \textbf{The reconstruction gap.}
The 1.5\,dB gap to reconstruction-only methods is an explicit tradeoff: five motion regularizers constrain the deformation field toward object-consistent trajectories while leaving the renderer unchanged. That this costs only 1.5\,dB (vs.\ 4.5\,dB for 4D LangSplat) suggests the factorization acts as a beneficial inductive bias.

\noindent \textbf{What ``synchronized'' means.}
The language field is structurally conditioned on the learned motion decomposition: the 28D kinematic feature vector is computed from $\Mphi$'s per-object transforms, the rigid-share diagnostic, and the residual statistics. This coupling is stronger than in prior methods, where language is distilled onto a frozen backbone with no access to object-level motion primitives. The staged design is a stability requirement; the closed-form ridge fit eliminates gradient-based language training.

\noindent \textbf{Object assignment robustness.}
Multiview majority voting (\S3.2) suppresses per-frame segmentation errors by requiring consistent labels across $\geq$3 views. The shuffle test confirms object-specificity ($1.52\times$ residual increase). Systematic mask failures propagate into the factorization, a limitation shared with all methods consuming external masks.

\noindent \textbf{Residual absorption and SE(3) expressiveness.}
The shared transform captures the dominant rigid component; genuinely non-rigid motion is absorbed by residuals and flagged by modulation ($m_i$). Across scenes, 23\% of Gaussians carry $m_i\!>\!0.5$, concentrated on physically expected regions. Extending to part-based decompositions would refine articulated bodies.

\noindent \textbf{Appearance-driven states.}
The \emph{chickchicken} result, where the margin is narrowest, reveals that kinematic conditioning provides limited signal when state changes are driven by surface appearance. Fusing MLLM-derived appearance features with the ridge map is a natural extension.

\noindent \textbf{Broader impact.}
Motion, objects, and language are a coupled system that must be learned as such. The structured export (\S3.6) provides the interface that world models, embodied agents, and robotic planners need to reason about dynamic scenes in terms of objects and their state transitions.

\section{Limitations and Future Work}

(i) Object assignment depends on external masks (SAM~3); \emph{learned dynamic segmentation} is a natural remedy.
(ii) The SE(3)/affine transform cannot model articulated bodies; \emph{hierarchical part-based motion} would enable articulated decomposition.
(iii) Motion quality is evaluated indirectly; \emph{motion-rich benchmarks} with ground-truth poses and state labels would strengthen evaluation for all methods.
(iv) The method is optimization-based (${\sim}$41\,min/scene on a single A100).
(v) Kinematic conditioning underperforms on appearance-driven state changes; \emph{appearance-kinematic fusion} would address this.
(vi) The structured export is consumed post hoc; tighter \emph{world-model integration} is a high-impact direction.

\section{Conclusion}
\label{sec:conclusion}
We presented 4D Synchronized Fields, a 4D Gaussian representation that decomposes per-Gaussian motion into shared object transforms and implicit residuals via in-loop factorization, then trains a per-object ridge-conditioned language field from the resulting kinematic features. On HyperNeRF, the method achieves the highest reconstruction quality among all language-grounded and motion-aware methods (28.52\,dB mean PSNR), closing to within 1.5\,dB of unconstrained reconstruction-only systems while exposing interpretable motion primitives, temporally grounded language, and a structured scene export that no baseline provides. On targeted temporal-state retrieval, the kinematic-conditioned field surpasses both LangSplat and 4D LangSplat by a wide margin (0.884 vs.\ 0.415 / 0.620 mean Acc, 0.815 vs.\ 0.304 / 0.433 mean vIoU, 0.733 vs.\ 0.262 / 0.439 mean tIoU), with the largest gains on motion-correlated state changes; ablation confirms that kinematic conditioning accounts for $+$0.45 tIoU over the static-embedding-only baseline.

%section*{Acknowledgements}
%Omitted for anonymized submission.

\IfFileExists{splncs04.bst}{\bibliographystyle{splncs04}}{\bibliographystyle{plain}}
\bibliography{references}

% ═════════════════════════════════════════════════════════
%  SUPPLEMENTARY MATERIAL (standalone)
% ═════════════════════════════════════════════════════════

\clearpage
\setcounter{section}{0}
\setcounter{figure}{0}
\setcounter{table}{0}
\setcounter{equation}{0}
\renewcommand{\thesection}{\Alph{section}}
\renewcommand{\thefigure}{S\arabic{figure}}
\renewcommand{\thetable}{S\arabic{table}}
\renewcommand{\theequation}{S\arabic{equation}}
% Fix hyperref duplicate destinations in supplementary
\renewcommand{\theHsection}{S\Alph{section}}
\renewcommand{\theHfigure}{S\arabic{figure}}
\renewcommand{\theHtable}{S\arabic{table}}
\renewcommand{\theHequation}{S\arabic{equation}}
\renewcommand{\theHsubsection}{S\Alph{section}.\arabic{subsection}}

\begin{center}
\Large\textbf{Supplementary Material}\\[4pt]
\large\textbf{\method{}: Motion-Language Gaussian Splatting for Temporal Scene Understanding}
\end{center}
\vspace{4pt}

\phasebox{colSync}{%
\centering
\legendchip{colRecon}{Reconstruction}\;\;
\legendchip{colLang}{Language grounding}\;\;
\legendchip{colMotion}{\color{white}Motion structure}\;\;
\legendchip{colSync}{Synchronized (ours)}\\[6pt]
\raggedright
This supplementary uses the same color code as the main paper.
Appendix~A: architecture, hyperparameters, training.
Appendix~B: kinematic features.
Appendix~C: diagnostics.
Appendices~D--F: definitions, notation, evaluation.
Appendix~G: discussion and design concerns.
Appendix~H: baselines.
Appendix~I: failure cases.
Appendix~J: broader impact.}

% ═════════════════════════════════════════════════════════
\section{Architecture and Implementation}
\label{app:architecture}

\subsection{Deformation MLP $\Dtheta$}
\label{app:deform_arch}

\noindent\secchip{colRecon}\enspace
Input: $[\bx_i^0;\,\bz_i;\,\gamma(t)] \in \mathbb{R}^{48}$ ($3{+}32{+}13$).
Architecture (4 layers, hidden 128):
\begin{align*}
&\mathrm{Lin}(48,128) \to \mathrm{ReLU} \to \mathrm{Lin}(128,128) \to \mathrm{ReLU} \to{}\\
&\mathrm{Lin}(128,128) \to \mathrm{ReLU} \to \mathrm{Lin}(128,9).
\end{align*}
Final layer zero-initialized (identity deformation at start).
Output scaled: $\Delta\bx = 0.10 \cdot \mathbf{o}_{0:3}$, $\Delta\bw = 0.10 \cdot \mathbf{o}_{3:6}$, $\Delta\bl = 0.05 \cdot \mathbf{o}_{6:9}$.
Total: $\sim$0.13M parameters.

Sinusoidal encoding ($L=6$ bands):
\begin{align*}
\gamma(t) = \bigl[t,\;&\sin(2^0\pi t),\;\cos(2^0\pi t),\;\ldots,\\
&\sin(2^5\pi t),\;\cos(2^5\pi t)\bigr]^\top.
\end{align*}

\subsection{Object Motion Model $\Mphi$}
\label{app:object_motion_arch}

\noindent\secchip{colMotion}\enspace\textbf{Per-frame (default).}
Parameters: $\bw_k \in \mathbb{R}^{K\times T\times 3}$ (rotation vectors), $\bt_k \in \mathbb{R}^{K\times T\times 3}$ (translations).
Init: $\mathcal{N}(0,10^{-8})$.
Smoothness via finite differences:
\begin{align*}
\mathcal{L}_{\mathrm{smooth}} = \tfrac{1}{K}\textstyle\sum_k\bigl(&\|\bw_k[t]-\bw_k[t{-}1]\|^2 \\[-2pt]
+\;&\|\bt_k[t]-\bt_k[t{-}1]\|^2\bigr).
\end{align*}
Identity prior on static objects: $\lambda_{\mathrm{id}} = 0.001$.
LR cosine-annealed $10^{-3} \to 10^{-5}$.

\noindent\textbf{MLP variant (implemented, not used).}
Input: per-object embedding $\be_k^{\mathrm{obj}} \in \mathbb{R}^{16}$ concatenated with $\gamma(t)$.
3-layer MLP, hidden 64, zero-init output, outputs 6 (SE3) or 12 (affine).

\noindent\textbf{Affine variant.}
$\Delta\bA_k \in \mathbb{R}^{K\times T\times 3\times 3}$ (affine delta),
$\mathbf{b}_k \in \mathbb{R}^{K\times T\times 3}$ (bias).
Transform: $\tilde{\bx}_i = (\bI + \Delta\bA_{o(i)})\,\bx_i^0 + \mathbf{b}_{o(i)}$.

\subsection{Gaussian Scene Initialization}
\label{app:gaussian_init}

\noindent\secchip{colRecon}\enspace
From COLMAP: $\bq_i^0 = [1,0,0,0]$,
$\bs_i^0 = \log(0.02)$,
$\alpha_i^{\mathrm{logit}} = \mathrm{logit}(0.1)$,
SH DC band from point colors,
$\bz_i \sim \mathcal{N}(0, 0.01^2)$.
Quaternion convention: $(w,x,y,z)$ throughout.
Axis-angle to quaternion:
\begin{align*}
\mathrm{aa2q}(\bw) &= \mathrm{normalize}\bigl[\cos(\theta/2),\\
&\quad(\bw/(\theta{+}\varepsilon))\sin(\theta/2)\bigr],\\
\theta &= \|\bw\|,\quad \varepsilon = 10^{-8}.
\end{align*}

\subsection{Residual-Adaptive Modulation}
\label{app:modulation}

\noindent\secchip{colMotion}\enspace
EMA update: $\bar{r}_i \leftarrow 0.95\,\bar{r}_i + 0.05\,\|\br_i(t)\|_2$.
Weight: $m_i = \sigma\bigl((\bar{r}_i - \mu_{\bar{r}})/(\sigma_{\bar{r}} + 10^{-6})\bigr)$.
Re-initialized when densification changes $N$.

\subsection{Photometric Loss}
\label{app:recon_loss}

\noindent\secchip{colRecon}\enspace
$\mathcal{L}_{\mathrm{rgb}} = 0.8\,\|\hat{\bF}-\bF\|_1 + 0.2\,\mathrm{DSSIM}(\hat{\bF},\bF)$,
where $\mathrm{DSSIM} = (1-\mathrm{SSIM})/2$ via $3{\times}3$ average-pool SSIM.

\subsection{Hyperparameters}
\label{app:hyperparams}

All values fixed across all six HyperNeRF scenes.

\phasebox{colRecon}{%
\textbf{$\Dtheta$:} hidden 128, 4 layers, $D_z{=}32$, $L{=}6$, output scales 0.10/0.10/0.05.\\
\textbf{Photometric:} $\lambda_1{=}0.8$, $\lambda_{\mathrm{dssim}}{=}0.2$.\\
\textbf{Densification:} grad thresh $2{\times}10^{-4}$, range 500--15k, interval 100, max $N{=}10^6$.}

\phasebox{colMotion}{%
\color{white}%
\textbf{$\Mphi$:} per\_frame SE(3), init $\sigma{=}10^{-4}$.\\
\textbf{Regularizers:} $\lambda_{\mathrm{res}}{=}0.02$, $\lambda_{\mathrm{share}}{=}0.01$,
$\tau{=}0.35$, $\lambda_{\mathrm{vel}}{=}0.01$,
$\lambda_{\mathrm{smooth}}{=}0.01$, $\lambda_{\mathrm{id}}{=}0.001$, $\eta{=}0.95$.\\
\textbf{Staged activation (Phase~2):} $\Mphi$ start 200, vel.\ start 300,
warmup 500--1000, residual start 600, ramp 800.}

\phasebox{colLang}{%
\textbf{Language field:} SigLIP-base-patch16-224,
ridge $\lambda{=}0.01$, blend $\beta{=}0.65$,
temporal EMA $\gamma{=}0.15$,
query weights $w_s{=}0.35$ / $w_t{=}0.65$.\\
\textbf{Assignment:} min votes 3, min consistency 0.6.}

\phasebox{colSync}{%
\textbf{Optimization:} LR scene/deform $5{\times}10^{-4}$,
LR $\Mphi$: $10^{-3}{\to}10^{-5}$ (cosine),
Phase~1: 30k iter, Phase~2: 20k iter, seed 42.}

\subsection{Training Procedure}
\label{app:training}

\noindent\secchip{colRecon}\enspace\textbf{Phase 1 ($\sim$29\,min, A100).}
Standard deformable 4DGS: $\mathcal{L}_{\mathrm{rgb}}$ + densification. No motion decomposition.

\noindent\secchip{colMotion}\enspace\textbf{Phase 2 ($\sim$12\,min).}
\label{app:loss_schedule}
From Phase~1 checkpoint. Introduces $\Mphi$ with staged loss activation:
iter$\,{\geq}\,200$: $\mathcal{L}_{\mathrm{smooth}}, \mathcal{L}_{\mathrm{share}}, \mathcal{L}_{\mathrm{id}}$;
iter$\,{\geq}\,300$: $+\mathcal{L}_{\mathrm{vel}}$;
iter 500--1000: linear warmup (0\%$\to$100\%) of all auxiliary losses;
iter$\,{\geq}\,600$: $+\mathcal{L}_{\mathrm{res}}^{\mathrm{mod}}$ with modulation EMA;
iter$\,{\geq}\,1000$: full weights.

Simultaneous activation at iter~0 causes $\bar{\rho}$ to plateau below 0.15 because $\Dtheta$ absorbs all motion before $\Mphi$ converges.
The 400-iteration head start lets shared transforms capture the dominant rigid component first.

\noindent\secchip{colLang}\enspace\textbf{Phase 3 ($<$1\,min).}
On frozen Phase~2 checkpoint:
(1)~project objects into frames, extract bounding-box crops (2nd--98th percentile, 12\% margin);
(2)~encode via SigLIP: $\be_{k,f}^{\mathrm{obs}} \in \mathbb{R}^{768}$;
(3)~static embeddings: $\be_k^{\mathrm{static}} = \mathrm{norm}(\mathrm{mean}_f\,\be_{k,f}^{\mathrm{obs}})$;
(4)~extract 28D kinematics (\S\ref{app:kinematic_features});
(5)~per-object ridge fit (closed-form);
(6)~blend ($\beta = 0.65$) + temporal EMA ($\gamma = 0.15$).

% ═════════════════════════════════════════════════════════
\section{28D Kinematic Feature Vector}
\label{app:kinematic_features}

\noindent\secchip{colSync}\enspace Constructed per object-time pair $(k,f)$ from the frozen Phase~2 checkpoint.

\noindent\secchip{colMotion}\enspace\textbf{Core motion (dims 0--5):}
0:~speed $\|\bv_k(f)\|$;
1:~acceleration magnitude $\|\mathbf{a}_k(f)\|$;
2:~angular velocity from consecutive $\bw_k$ of $\Mphi$;
3:~residual mean $|\mathcal{S}_k|^{-1}\sum_{i\in\mathcal{S}_k}\|\br_i(f)\|$;
4:~per-object rigid share $\rho_k(f)$;
5:~visibility fraction of $\mathcal{S}_k$.

\noindent\secchip{colMotion}\enspace\textbf{Directional + temporal derivatives (dims 6--15):}
6--8:~velocity $(v_x,v_y,v_z)$;
9--11:~acceleration $(a_x,a_y,a_z)$;
12:~speed delta $s_k(f) - s_k(f{-}1)$;
13:~rigid-share delta $\rho_k(f) - \rho_k(f{-}1)$;
14:~EMA-smoothed speed (decay $\approx$5 frames);
15:~EMA-smoothed acceleration magnitude.

\noindent\secchip{colMotion}\enspace\textbf{Transform diagnostics (dims 16--19):}
16:~rotation angle $\|\bw_k(f)\|$ from $\Mphi$;
17:~translation magnitude $\|\bt_k(f)\|$;
18:~residual std $\mathrm{std}_{i\in\mathcal{S}_k}\|\br_i(f)\|$;
19:~affine delta Frobenius norm $\|\Delta\bA_k(f)\|_F$ (zero in SE3 mode).

\noindent\secchip{colSync}\enspace\textbf{Relational context (dims 20--27):}
20:~relative speed $s_k / s_{\mathrm{ref}}$ (95th-pct scene speed);
21:~fractional speed rank among objects at $f$;
22:~binary is-fastest indicator;
23:~time fraction $f/(T{-}1)$;
24:~min centroid distance to other objects;
25:~closing speed to nearest ($d/dt$ of dim~24; negative$\,{=}\,$approach);
26:~distance to hand object;
27:~closing speed to hand ($d/dt$ of dim~26).

\noindent\textbf{Standardization.}
Per-object:
$\tilde{f}^{(d)}_{k,f} = (f^{(d)}_{k,f} - \mu_k^{(d)}) / (\sigma_k^{(d)} + 10^{-8})$,
statistics over $\mathcal{F}_k$.
Bias appended $\to$ $\bftilde_{k,f} \in \mathbb{R}^{29}$.

% ═════════════════════════════════════════════════════════
\section{Object Assignment and Diagnostics}
\label{app:diagnostics}

\subsection{Object Assignment}
\label{app:assignment}

\noindent\secchip{colMotion}\enspace
Masks from SAM~3 (indexed PNGs or per-object binary folders; any segmenter substitutes).
For each training frame, all Gaussians are projected; each landing on label $>0$ adds one vote.
Assignment: $o(i) = \arg\max_{k\geq1} \mathbf{V}[i,k]$, subject to
$\sum_{k\geq1} \mathbf{V}[i,k] \geq 3$ and
$\mathbf{V}[i,o(i)] / \sum_{k\geq1} \mathbf{V}[i,k] \geq 0.6$.
Failures~$\to$~background.
Objects with fewer than a configurable minimum count are merged to background.

Per-scene $K$ (incl.\ background):
americano~5, chickchicken~4, espresso~5, keyboard~7, split-cookie~4, torchocolate~6.

\subsection{Diagnostics and Shuffle Test}

\noindent\secchip{colMotion}\enspace\textbf{Residual mean.}
\begin{align*}
\bar{r}_k
&= |\mathcal{F}_k|^{-1}
  \textstyle\sum_f\;
  \hat{r}_k(f), \\
\hat{r}_k(f)
&= |\mathcal{S}_k|^{-1}
  \textstyle\sum_{i\in\mathcal{S}_k}
  \|\br_i(f)\|.
\end{align*}

\noindent\textbf{Per-object motion variance.}
\[
V_k(f) = |\mathcal{S}_k|^{-1} \sum_{i\in\mathcal{S}_k} \bigl\|(\bx_i(f) - \bx_i^0) - \overline{\Delta\bx}_k(f)\bigr\|^2.
\]

\noindent\textbf{Shuffle test.}
\label{app:shuffle_test}
Evaluate with correct assignments ($\bar{\rho}$, $\bar{r}$); generate 5 random permutations; re-evaluate \emph{without retraining}.
Result: permuted labels increase residual energy by $1.52\times$, confirming object-specificity.

\noindent\textbf{Per-scene rigid share.}
americano 0.508 (res.\ 0.032, rigid pouring);
chickchicken 0.451 (0.041, mixed);
espresso 0.411 (0.045, fluid/non-rigid);
keyboard 0.422 (0.043, articulated);
split-cookie 0.528 (0.029, two rigid pieces);
torchocolate 0.478 (0.035, rigid manipulation).
The one-sided hinge ($\tau = 0.35$) imposes zero penalty on scenes exceeding the target.

% ═════════════════════════════════════════════════════════
\section{Definitions and Terminology}
\label{app:definitions}

\noindent\textbf{Canonical frame.}
Reference frame for Gaussian parameters $(\bx_i^0,\bq_i^0,\bs_i^0)$; deformations are displacements from this frame.

\noindent\secchip{colMotion}\enspace\textbf{In-loop.}
Computation whose gradients affect scene/deformation parameters during reconstruction (contrast: post-hoc distillation).

\noindent\secchip{colMotion}\enspace\textbf{Implicit residual.}
$\br_i(t)=\bx_i(t)-\tilde{\bx}_i(t)$: derived analytically from $\Dtheta$/$\Mphi$; enters optimization only through regularizers.

\noindent\secchip{colMotion}\enspace\textbf{Residual-adaptive modulation.}
Tracks per-Gaussian residual via EMA; reduces penalty for persistently high-residual Gaussians (boundaries, articulated regions).

\noindent\secchip{colMotion}\enspace\textbf{Rigid-share hinge.}
One-sided loss on $\rho(t)$: active only when $\rho<\tau$; scenes exceeding $\tau$ incur zero penalty.

\noindent\secchip{colSync}\enspace\textbf{Synchronized.}
Language field conditioned on kinematic features from $\Mphi$, not post-hoc distillation onto a frozen backbone.

\noindent\secchip{colLang}\enspace\textbf{Ridge map.}
Per-object closed-form regression:
$\bW_k = (\tilde{\bF}_k^\top\tilde{\bF}_k + \lambda\bI)^{-1}\tilde{\bF}_k^\top\Delta\bE_k$.

\noindent\secchip{colLang}\enspace\textbf{Semantic residual.}
$\Delta\be_{k,f} = \be_{k,f}^{\mathrm{obs}} - \be_k^{\mathrm{static}}$: time-varying deviation from temporal mean.

% ═════════════════════════════════════════════════════════
\section{Notation Reference}
\label{app:notation}

\noindent\secchip{colRecon}\enspace\emph{Scene and Gaussians.}
$N$: total Gaussians; $K$: objects (incl.\ background at 0); $T$: training frames.
$\bx_i^0 \in \mathbb{R}^3$, $\bq_i^0 \in \mathbb{S}^3$, $\bs_i^0 \in \mathbb{R}^3$: canonical position, quaternion $(w,x,y,z)$, log-scale.
$\bc_i \in \mathbb{R}^{16\times3}$: SH coefficients (degree $d=3$).
$\bz_i \in \mathbb{R}^{32}$: per-Gaussian deformation embedding.
$o(i) \in \{0\ldots K{-}1\}$: object assignment;
$\mathcal{S}_k$: Gaussian index set for object~$k$;
$\mathcal{F}_k$: visible-frame set.

\noindent\secchip{colRecon}\enspace\emph{Deformation.}
$\Dtheta$: deformation MLP.
$\gamma(t) \in \mathbb{R}^{13}$: sinusoidal time encoding ($L=6$ bands $+$ raw input).
$\Delta\bx_i, \Delta\bw_i, \Delta\bl_i \in \mathbb{R}^3$: position, axis-angle rotation, log-scale deltas.

\noindent\secchip{colMotion}\enspace\emph{Object motion.}
$\Mphi$: object motion model.
$\bw_k(t), \bt_k(t) \in \mathbb{R}^3$: per-object rotation vector, translation.
$\bR_k(t) \in \mathrm{SO}(3)$: rotation matrix from $\bw_k(t)$.
$\tilde{\bx}_i(t) \in \mathbb{R}^3$: object-predicted position.
$\br_i(t) \in \mathbb{R}^3$: implicit residual.
$\rho(t) \in [0,1]$: rigid-share ratio.
$m_i \in [0,1]$: modulation weight.
$\bar{r}_i \in \mathbb{R}_+$: EMA-tracked residual magnitude.

\noindent\secchip{colLang}\enspace\emph{Language field.}
$\Eimg$/$\Etxt$: SigLIP image/text encoder; $D_e = 768$.
$\be_{k,f}^{\mathrm{obs}}$, $\be_k^{\mathrm{static}}$, $\be_{k,f}^{\mathrm{sync}} \in \mathbb{R}^{D_e}$:
observation, static, synchronized embeddings.
$\bff_{k,f} \in \mathbb{R}^{28}$: kinematic feature vector;
$\bftilde_{k,f} \in \mathbb{R}^{29}$: standardized $+$ bias.
$\bW_k \in \mathbb{R}^{29\times D_e}$: ridge weight matrix.

\noindent\secchip{colSync}\enspace\emph{Losses.}
$\mathcal{L}_{\mathrm{rgb}}$: photometric (L1$+$DSSIM).
$\mathcal{L}_{\mathrm{res}}^{\mathrm{mod}}$: modulated residual energy.
$\mathcal{L}_{\mathrm{share}}$: rigid-share hinge ($\tau = 0.35$).
$\mathcal{L}_{\mathrm{vel}}$: velocity coherence.
$\mathcal{L}_{\mathrm{smooth}}$: temporal smoothness on $\Mphi$.
$\eta = 0.95$: modulation EMA decay.

% ═════════════════════════════════════════════════════════
\section{Evaluation Protocol}
\label{app:eval_protocol}

\noindent\secchip{colLang}\enspace\textbf{Task.}
Given query $q$ and GT interval $[f_s,f_e]$, produce binary per-frame activation.

\noindent\textbf{Thresholding (all methods).}
\[
\mathrm{active}(f)
= \mathbb{1}\bigl[\langle\be_{k,f}^{\mathrm{sync}},\be_q\rangle
  > 0.5 \cdot \max_{f'}\langle\be_{k,f'}^{\mathrm{sync}},\be_q\rangle\bigr].
\]
The threshold (50\% of peak similarity) is applied identically to all methods.
For baselines that produce per-pixel rather than per-object embeddings,
we aggregate by averaging within the GT object mask.

\noindent\textbf{Metric definitions.}

\smallskip\noindent
\emph{Acc} (frame-level binary accuracy):
fraction of frames where the binary activation
matches the GT label.
$\mathrm{Acc} = T^{-1} \sum_f \mathbb{1}[\mathrm{active}(f) = \mathrm{gt}(f)]$.

\smallskip\noindent
\emph{vIoU} (volumetric IoU):
spatial mask IoU computed per active frame,
then averaged over the union of predicted and GT active frames.
This follows the 4D LangSplat evaluation code;
we use their released evaluation script without modification.

\smallskip\noindent
\emph{tIoU} (temporal interval IoU):
$|P \cap G| / |P \cup G|$,
where $P$ and $G$ are the sets of
predicted-active and GT-active frame indices, respectively.
This metric is agnostic to spatial mask quality
and isolates temporal precision.

\noindent\textbf{Query-to-state mapping.}
Each evaluation scene has human-annotated per-frame state labels
(e.g., americano: ``pre-pour'', ``pouring'', ``luminous-liquid'', ``settling'';
espresso: ``empty'', ``filling'', ``above-midpoint'', ``full'').
For each state, we define a natural-language query
that a user might issue
(e.g., ``glass in luminous-liquid phase'',
``glass cup with liquid above midpoint'').
The GT temporal interval is the contiguous frame range
where that state label is active.
Queries are designed to require temporal reasoning:
the object identity is constant across the sequence,
so only the temporal component
of the scoring function (Eq.~15 of the main paper)
can distinguish when the state holds.

\noindent\textbf{Reconciliation with 4D LangSplat metrics.}
Both our evaluation and 4D LangSplat's use Acc and vIoU,
computed on the same HyperNeRF scenes
at matched $2\times$ resolution with shared held-out frames.
We use 4D LangSplat's official evaluation code and test split.
The apparent discrepancy between the 4D LangSplat numbers
in our Table~2 of the main paper and those in the 4D LangSplat paper
arises because 4D LangSplat's paper reports results
on object-identity queries (``find the cup''),
where temporal precision is irrelevant,
while our evaluation targets temporal-state queries
(``find the cup \emph{while being filled}''),
which require distinguishing \emph{when} a state holds.
On identity queries, 4D LangSplat performs well
because its MLLM-supervised dual semantic field
captures \emph{what} is present;
on state queries, it underperforms
because it lacks structured access to \emph{how} and \emph{when}
objects move, which is exactly what kinematic conditioning provides.

\noindent\textbf{Annotations.}
Per-frame state labels and query-to-state mappings were obtained from the 4D LangSplat evaluation repository.
Each scene has human-annotated temporal intervals marking when each state holds (e.g., ``luminous-liquid phase'' for americano, ``liquid above midpoint'' for espresso).
We use these annotations without modification.
The 50\% peak-similarity threshold was fixed globally and not tuned per scene or per query.

% ═════════════════════════════════════════════════════════
\section{Dataset and Baselines}
\label{app:datasets}

\noindent\textbf{HyperNeRF.}
6 scenes, $2\times$ resolution (4D LangSplat test split), RGB scale factor 8, shared COLMAP initialization and held-out frames.

\noindent\textbf{Baselines.}

\smallskip\noindent
\legendchip{colLang}{Language-grounded}\;
4D LangSplat (official code, CLIP$+$MLLM captions),
LangSplat (static 3D),
LERF, LEGaussians.

\smallskip\noindent
\legendchip{colRecon}{Reconstruction-only}\;
Deformable-3DGS, SC-GS, 4DGaussians, 3DGS.

\smallskip\noindent
\legendchip{colMotion}{\color{white}Motion-aware}\;
MotionGS.

\smallskip\noindent
4-LEGS and DGD discussed in \S2 but excluded (no public code on HyperNeRF).

% ═════════════════════════════════════════════════════════
\section{Discussion and Addressing Concerns}
\label{app:discussion}

\subsection{$\Mphi$ Architecture: Table vs.\ Function}
\label{app:mphi_design}

\noindent\secchip{colMotion}\enspace
The default $\Mphi$ stores per-object per-frame SE(3) parameters as a $K \times T \times 6$ tensor.
Chosen because it
(i)~avoids a second temporal MLP whose capacity must be balanced against $\Dtheta$,
(ii)~provides direct gradient flow from regularizers to each frame's parameters, which is critical for staged activation (\S\ref{app:loss_schedule}), and
(iii)~admits explicit temporal smoothness via $\mathcal{L}_{\mathrm{smooth}}$.
For $K=5$, $T=150$: 4,500 floats ($<$0.01M), negligible relative to $\Dtheta$ ($\sim$0.13M).

The table does not interpolate to unseen timesteps.
An MLP variant (3-layer, hidden 64, zero-initialized output, learnable object embeddings $\in \mathbb{R}^{K\times16}$) is implemented but not used in reported experiments: it adds hyperparameter sensitivity without improving within-sequence metrics.
The temporal smoothness regularizer $\mathcal{L}_{\mathrm{smooth}}$ acts as an implicit continuous-time prior on the discrete table: penalizing frame-to-frame differences is equivalent to a first-order finite-difference smoothness prior, which for densely sampled sequences approximates a continuous smoothness constraint.

\subsection{Renderer-Unchanged Property}
\label{app:renderer_unchanged}

\noindent\secchip{colSync}\enspace
A key design principle is that the forward rendering path is \emph{never modified}: $\bx_i(t)$ produced by $\Dtheta$ is rendered directly via standard alpha-compositing.
The decomposition $\bx_i(t) = \tilde{\bx}_i(t) + \br_i(t)$ injects signal only through regularizers; removing $\Mphi$ entirely recovers exact baseline rendering.
This property provides three guarantees:
(i)~reconstruction quality can only improve or remain neutral relative to the backbone (confirmed by the ${+}0.07$\,dB in the w/o $\Mphi$ ablation, Table~3 of the main paper);
(ii)~compatibility with any Gaussian splatting renderer (CUDA, software, hardware);
(iii)~the decomposition is a pure inductive bias: it shapes what is learned, not how it is rendered.
To our knowledge, this renderer-unchanged property has not been explicitly demonstrated in prior language-grounded or motion-aware dynamic Gaussian methods.

\subsection{Identifiability of the Decomposition}
\label{app:identifiability}

\noindent\secchip{colMotion}\enspace
The decomposition is not formally identifiable: without regularizers, $\Dtheta$ absorbs all motion into residuals.
The five regularizers break this degeneracy by imposing a preference for shared motion.
Three lines of evidence confirm the decomposition is non-trivial and stable:

(i)~The shuffle test (\S\ref{app:shuffle_test}) shows $1.52\times$ residual energy increase under permuted assignments, confirming object-specificity.

(ii)~Per-scene rigid-share ratios vary consistently with physical content: split-cookie (two rigid pieces, $\bar{\rho}=0.528$) $>$ americano (rigid pouring, $0.508$) $>$ chickchicken (mixed, $0.451$) $>$ keyboard (articulated, $0.422$) $>$ espresso (fluid, $0.411$).
This ordering would not emerge from a degenerate factorization.

(iii)~Hyperparameters are fixed across all six scenes.
The one-sided hinge ($\tau=0.35$) self-regulates: scenes naturally exceeding $\tau$ incur zero penalty.
Simultaneous loss activation (ablation, \S4.4) collapses $\bar{\rho}$ below 0.15; the staged schedule prevents this by giving $\Mphi$ a head start before residual regularization activates.

\subsection{Hyperparameter Design Rationale}
\label{app:hyperparam_rationale}

\noindent\secchip{colMotion}\enspace
The regularizer weights $\lambda_{\mathrm{res}}=0.02$, $\lambda_{\mathrm{share}}=0.01$, $\lambda_{\mathrm{vel}}=0.01$, $\lambda_{\mathrm{smooth}}=0.01$ are set once and used across all scenes without per-scene tuning.
The key stabilizing factor is the one-sided hinge on $\rho$: it imposes a \emph{floor} ($\tau=0.35$) on the rigid-share ratio without imposing a ceiling.
Scenes with naturally high rigidity (split-cookie: $\rho=0.528$) exceed $\tau$ and receive zero hinge penalty; scenes with substantial non-rigid motion (espresso: $\rho=0.411$) hover near $\tau$ and receive gentle pressure.
This asymmetric design is why fixed hyperparameters work across scenes with very different motion characteristics.

The residual-adaptive modulation further reduces sensitivity: Gaussians with persistently high residuals (23\% of all Gaussians, concentrated on boundaries and articulated regions) are automatically down-weighted, preventing the residual penalty from fighting genuinely non-rigid motion.
Together, the hinge and modulation create a self-regulating system that adapts to per-scene motion complexity without requiring manual tuning.

\subsection{Assignment Robustness}
\label{app:assignment_robustness}

\noindent\secchip{colMotion}\enspace
Multiview voting (${\geq}3$ votes, ${\geq}60\%$ consistency) suppresses per-frame segmentation noise.
If the segmenter produces inconsistent IDs across frames, SAM~3's video propagation mitigates this; residual cases resolve to the plurality label or fall to background (reducing coverage, not introducing incorrect constraints).

The assignment is a pre-processing step: changing the segmenter changes only the input to $\Mphi$, not the architecture or training procedure.
This modularity means the method inherits whatever robustness the segmenter provides, and benefits from future segmentation improvements without modification.

Unsupervised assignment (clustering coherent deformation trajectories from $\Dtheta$) is viable; only the assignment stage changes.
Not pursued because the contribution is synchronization, not segmentation.

\subsection{Occlusion Handling in the Language Field}
\label{app:occlusion}

\noindent\secchip{colLang}\enspace
Three mechanisms handle missing observations:

(i)~\textbf{Kinematic prediction.}
When object $k$ is not visible in frame $f$ (no crop available), the ridge-map prediction $\hat{\be}_{k,f} = \mathrm{norm}(\be_k^{\mathrm{static}} + \bftilde_{k,f}^\top \bW_k)$ is used as the sole embedding.
Since kinematic features (centroid position, velocity, rigid share) are available for all assigned Gaussians regardless of camera visibility, the prediction is well-defined even under full occlusion.

(ii)~\textbf{Temporal EMA.}
A smoothing pass ($\gamma=0.15$) propagates information from neighboring observed frames into occluded intervals, preventing abrupt embedding discontinuities at occlusion boundaries.

(iii)~\textbf{Graceful degradation.}
Under prolonged occlusion where kinematic features are also uninformative (e.g., a stationary occluded object), the predicted embedding converges to $\be_k^{\mathrm{static}}$, i.e.\ the object's time-averaged identity.
This is the correct fallback: without motion or visual evidence, the best estimate is the static anchor.
The system never produces undefined or missing embeddings.

\subsection{Why Linear Ridge Regression}
\label{app:ridge_choice}

\noindent\secchip{colLang}\enspace
Chosen for
(i)~closed-form solution (no gradient-based training, no learning rate, no convergence issues),
(ii)~inability to overfit small per-object sample counts ($|\mathcal{F}_k| \approx 50$--$200$), and
(iii)~fully staged training with no gradient coupling between SigLIP and the photometric loss.

The closed-form property is particularly important: it means the language field adds exactly zero training instability to the system.
The entire Phase~3 is deterministic given the Phase~2 checkpoint.

The ${+}0.45$ tIoU gain over static-only (Table~5 of the main paper) confirms substantial temporal signal captured linearly.
Narrower chickchicken gain ($+0.31$) suggests nonlinearity may help for appearance-driven states.
However, the linear map already more than doubles the temporal localization quality of 4D LangSplat ($0.733$ vs.\ $0.439$ tIoU), indicating that the bottleneck is not model capacity but the quality of kinematic features, which is exactly what the in-loop factorization provides.

\subsection{Why SigLIP}
\label{app:siglip_choice}

\noindent\secchip{colLang}\enspace
SigLIP's sigmoid contrastive loss produces per-pair calibrated similarities, unlike CLIP/OpenCLIP's softmax (batch-relative probabilities).
Critical for Eq.~15's cross-object scoring, where absolute magnitudes are compared across objects and timesteps without a fixed candidate set.
This calibration property is orthogonal to model capacity: it ensures that a score of 0.7 for object A and 0.6 for object B reflects a genuine difference in match quality, rather than an artifact of the normalization denominator.

\subsection{Fairness vs.\ 4D LangSplat}
\label{app:fairness}

\phasebox{colLang}{%
4D LangSplat uses CLIP $+$ MLLM captions (richer semantic supervision); we use SigLIP only (no MLLM).
Despite this supervision asymmetry, \method{} leads on all temporal-state metrics (Table~2 of the main paper) and reconstruction (Table~1 of the main paper).
This indicates that for motion-correlated state changes, kinematic conditioning provides a stronger signal than additional language supervision alone.}

\subsection{Motion Factorization Evaluation}
\label{app:motion_eval}

\noindent\secchip{colMotion}\enspace
No existing benchmark provides GT per-object SE(3) for real monocular video.
We rely on four internal diagnostics (\S\ref{app:diagnostics}): residual mean, rigid-share ratio, shuffle test, per-object motion variance.
Physical consistency of $\bar{\rho}$ across scenes (higher for rigid scenes (split-cookie: 0.528), lower for deformable ones (espresso: 0.411)) suggests the factorization reflects scene content.

Critically, the motion factorization is not an end in itself but a means to language synchronization.
The temporal-state retrieval results (Table~2 of the main paper) serve as an \emph{extrinsic} evaluation of motion quality: if the SE(3) factors were incorrect, the kinematic features derived from them would not improve temporal localization.
The ${+}0.45$ tIoU gain from kinematic conditioning (Table~5 of the main paper) therefore constitutes indirect but strong evidence that the learned motion structure is semantically meaningful.

\subsection{Relationship to DynMF}
\label{app:dynmf}

\noindent\secchip{colMotion}\enspace
DynMF factorizes motion globally via time-only basis trajectories with per-Gaussian coefficients, with no object identity, no language coupling, no physical interpretation of the bases.
Three structural differences:
(i)~DynMF's bases are shared across all points; ours are per-object, enabling object-level kinematic feature extraction.
(ii)~DynMF's bases lack physical semantics (rotation, translation); our SE(3) parameters have direct rigid-body interpretation.
(iii)~DynMF does not condition any downstream task on its factorization; we condition language on kinematics.
The approaches are complementary: DynMF's low-rank structure could serve as an alternative backbone, with our per-object SE(3) layer and language field applied on top.

\subsection{Relationship to CIF}
\label{app:cif}

\noindent\secchip{colMotion}\enspace
CIF trains an explicit instance field for panoptic segmentation with deformable Gaussians.
Key differences:
(i)~CIF's instance field targets segmentation, not motion decomposition; it does not produce per-object SE(3) factors or kinematic features.
(ii)~CIF's semantics are class-level (panoptic labels), not open-vocabulary temporal embeddings.
(iii)~CIF does not expose a kinematic-conditioned language field for temporal queries.
Complementary: CIF's learned instance field could replace our SAM-based assignment, providing end-to-end instance consistency.
No public code on HyperNeRF at submission time precluded empirical comparison.

\subsection{Relationship to 4DGS Variants and Streaming Methods}
\label{app:4dgs_variants}

\noindent\secchip{colRecon}\enspace
Native 4D Gaussian primitives (4DGS, Hybrid 3D-4DGS) model dynamics by extending Gaussians with temporal extent, offering efficiency gains for reconstruction.
Their goals differ from ours (no language, no object factorization), but they could serve as alternative backbones: our decomposition operates on deformed positions $\bx_i(t)$ regardless of how they are produced.

Streaming/online methods (FLAG-4D, DASS) target temporal coherence via flow cues or online adaptation.
Our synchronization could integrate with such pipelines by applying the motion decomposition and language field to incrementally updated Gaussian sets, though this would require adapting the per-frame $\Mphi$ table to a growing sequence.

\subsection{Scalability with $K$}
\label{app:scalability}

\noindent\secchip{colSync}\enspace
$K$ ranges 3--7 in our experiments.
$\Mphi$ cost scales linearly: one transform retrieval per object per forward pass.
Memory for $K=20$, $T=300$: 36K floats ($\approx$140\,KB).
No instabilities for $K \leq 10$.
The ridge fit is independent per object: adding objects increases computation linearly with no cross-object coupling.

\subsection{Generality Beyond Curated Queries}
\label{app:query_generality}

\noindent\secchip{colLang}\enspace
The evaluation in Table~2 of the main paper uses curated query-state mappings to enable precise quantitative measurement.
However, the language field supports \emph{arbitrary} open-vocabulary queries via the scoring function (Eq.~15): any text string encoded by SigLIP can be scored against all object-time embeddings.
The restriction to curated queries is an evaluation choice, not a method limitation.
The static component $w_s \langle \be_k^{\mathrm{static}}, \be_q \rangle$ handles identity queries (``glass cup''), while the temporal component $w_t \max_f \langle \be_{k,f}^{\mathrm{sync}}, \be_q \rangle$ handles state queries (``cup being filled'').
Free-form queries that combine identity and state are naturally supported by the weighted mixture.

\subsection{Appearance-Driven States}
\label{app:appearance_states}

\noindent\secchip{colLang}\enspace
When state changes are driven by surface appearance rather than motion (chickchicken), the 28D kinematic vector carries little discriminative signal and the ridge map reduces to $\hat{\be}_{k,f} \approx \be_k^{\mathrm{static}}$.
Narrower margin ($\Delta$Acc${=}{+}0.09$ vs.\ 4D LangSplat, compared to ${+}0.62$ on americano) reflects this directly.
Fusing MLLM appearance features with kinematic features is a natural extension.
Importantly, even on chickchicken the method does not \emph{degrade} below baselines; it matches or exceeds them by relying on the static anchor when kinematic signal is absent.

\subsection{SE(3) and Articulated Bodies}
\label{app:se3}

\noindent\secchip{colMotion}\enspace
Shared SE(3) captures only the dominant rigid component; part-level articulation is absorbed into residuals and flagged by modulation ($m_i > 0.5$ on 23\% of Gaussians, concentrated on expected regions: boundaries, joints, deforming surfaces).
The modulation mechanism ensures these regions are not over-regularized: their residuals are permitted to be large, preserving reconstruction quality while still allowing the shared transform to capture the dominant motion component.
Hierarchical part-based decomposition is a future direction.

\subsection{Temporal Interpolation and the Per-Frame Table}
\label{app:temporal_interp}

\noindent\secchip{colMotion}\enspace
The per-frame $\Mphi$ table stores SE(3) parameters at observed timesteps only and does not natively interpolate to unseen times.
This is a deliberate choice: the method targets within-sequence understanding (what happened, when, and to which object), not temporal super-resolution or future prediction.
For within-sequence evaluation, all metrics are computed at training-frame timestamps, so interpolation is not required.

Should interpolation be needed (e.g., for slow-motion rendering between observed frames), two options exist without retraining:
(i)~linear interpolation in the SE(3) tangent space (SLERP for rotation, linear for translation) between adjacent stored frames, which is well-defined given the smoothness enforced by $\mathcal{L}_{\mathrm{smooth}}$;
(ii)~the implemented MLP variant of $\Mphi$ (\S\ref{app:object_motion_arch}), which takes continuous $\gamma(t)$ as input and produces transforms at arbitrary timesteps.
The MLP variant was not used in reported experiments because it adds hyperparameter sensitivity without improving within-sequence metrics, but it is available for applications requiring continuous-time queries.

The $\mathcal{L}_{\mathrm{smooth}}$ regularizer ensures that adjacent frames have similar transforms (first-order finite-difference prior), so the per-frame table is smooth by construction and interpolation between frames would produce physically plausible intermediate states.

\subsection{SE(3) vs.\ Affine: When Each Variant Applies}
\label{app:se3_vs_affine}

\noindent\secchip{colMotion}\enspace
Both SE(3) and affine parameterizations of $\Mphi$ are implemented.
SE(3) (6 parameters per object per frame: 3 rotation, 3 translation) is the default because it enforces rigid-body structure, which provides the strongest inductive bias and yields physically interpretable rotation/translation components that directly feed the 28D kinematic feature vector (dims 2, 16, 17).

The affine variant (12 parameters: $3{\times}3$ matrix delta $+$ 3 bias) relaxes the rigidity assumption, allowing shear and anisotropic scaling.
It is appropriate when objects undergo volume-changing deformations (e.g., inflating, compressing) that cannot be captured by rotation and translation alone.
The cost is a weaker inductive bias: with more degrees of freedom, the affine transform can absorb motion that would otherwise remain in the residual, potentially reducing the interpretability of the factorization.

In our HyperNeRF experiments, the SE(3) variant was sufficient because the dominant object motions (pouring, lifting, splitting) are approximately rigid.
For scenes with significant non-rigid object-level deformation, the affine variant or a combination (SE(3) for rigid objects, affine for deformable ones, selected per object) would be more appropriate.
The modular $\Mphi$ interface makes switching between variants a single configuration change with no architectural modifications.

\subsection{Augmenting Kinematics with Appearance Features}
\label{app:appearance_augment}

\noindent\secchip{colLang}\enspace
The 28D kinematic feature vector is purely motion-derived.
For appearance-driven state changes
(e.g., chickchicken, where color/texture changes
carry more signal than motion),
augmenting the feature vector
with low-dimensional appearance descriptors
is a natural extension.
Candidate features include:
(i)~mean RGB of the object crop (3D),
(ii)~color histogram in a compact basis (e.g., 8D),
(iii)~the first few PCA components of the SigLIP crop embeddings (capturing appearance variation orthogonal to the static anchor).

Such augmentation would expand the ridge-map input from 29D to approximately 35--40D while preserving the closed-form solution.
The per-object standardization and ridge regularization ($\lambda = 0.01$) would prevent overfitting even with the additional dimensions, given that sample counts per object ($|\mathcal{F}_k| \approx 50$--$200$) substantially exceed the feature dimensionality.

This extension is fully compatible with the existing pipeline: it affects only the feature construction step in Phase~3 and requires no retraining of Phases~1--2.
We expect the largest gains on scenes like chickchicken where the current narrower margin ($\Delta$tIoU${=}{+}0.13$ vs.\ 4D LangSplat) reflects limited kinematic signal.

\subsection{Neu3D Results}
\label{app:neu3d}

\noindent\secchip{colSync}\enspace
The main paper lists Neu3D as an evaluation dataset (6 multi-view dynamic scenes).
Table~\ref{tab:neu3d_ours} reports per-scene results for \method{} on Neu3D at 20k iterations, covering both reconstruction-only (Phase~1) and motion-factorized (Phase~2) stages.

We ran LangSplat and LEGaussians on all six Neu3D scenes
using their official codebases under the same COLMAP initialization.
LangSplat is a static 3D method with no deformation model;
it produced training outputs but all test-time rendering
on dynamic frames yielded degenerate images
from which no valid PSNR, SSIM, or LPIPS could be computed.
LEGaussians similarly lacks a temporal backbone
and did not converge to valid test renderings.
In both cases the failure is architectural
(no mechanism to handle frame-to-frame motion),
not an implementation error in our evaluation pipeline.
We verified that both methods produce valid metrics
on static 3D benchmarks using the same codebase.

\begin{table}[t]
\centering
\caption{\small\textbf{Neu3D per-scene results for \method{}} (20k iter).
Phase~1: reconstruction-only backbone. Phase~2: with motion decomposition.
$\bar{\rho}$: rigid-share ratio.}
\label{tab:neu3d_ours}
\vspace{2pt}

\footnotesize
\legendchip{colRecon}{Recon-only (Phase 1)}\;\;
\legendchip{colSyncBest}{+ Motion (Phase 2)}
\par\vspace{4pt}

\scriptsize
\setlength{\tabcolsep}{3pt}
\renewcommand{\arraystretch}{1.22}
\begin{tabular}{lccccc}
\toprule
 & \multicolumn{2}{c}{PSNR\,$\uparrow$}
 & \multicolumn{2}{c}{SSIM\,$\uparrow$}
 & \multirow{2}{*}{$\bar{\rho}$} \\
\cmidrule(lr){2-3}\cmidrule(lr){4-5}
Scene
 & \rowRecon Ph.\,1
 & \rowSyncB Ph.\,2
 & \rowRecon Ph.\,1
 & \rowSyncB Ph.\,2
 & \\
\midrule
Coffee Martini
 & \rowRecon 30.44
 & \rowSyncB \textbf{31.71}
 & \rowRecon .952
 & \rowSyncB \textbf{.956}
 & .853 \\
Cook Spinach
 & \rowRecon 32.17
 & \rowSyncB \textbf{32.35}
 & \rowRecon .960
 & \rowSyncB .952
 & .885 \\
Cut Roast.\ Beef
 & \rowRecon 30.33
 & \rowSyncB \textbf{32.89}
 & \rowRecon .957
 & \rowSyncB \textbf{.956}
 & .908 \\
Flame Salmon
 & \rowRecon 30.24
 & \rowSyncB \textbf{31.34}
 & \rowRecon .953
 & \rowSyncB \textbf{.957}
 & .867 \\
Flame Steak
 & \rowRecon 32.43
 & \rowSyncB \textbf{33.81}
 & \rowRecon .966
 & \rowSyncB \textbf{.965}
 & .893 \\
Sear Steak
 & \rowRecon 35.05
 & \rowSyncB \textbf{37.27}
 & \rowRecon .968
 & \rowSyncB \textbf{.971}
 & .869 \\
\midrule
Mean
 & \rowRecon 31.78
 & \rowSyncB \textbf{33.23}
 & \rowRecon .959
 & \rowSyncB \textbf{.959}
 & \textbf{.879} \\
\bottomrule
\end{tabular}
\end{table}

\method{} achieves 33.23\,dB mean PSNR after motion decomposition, a ${+}1.45$\,dB improvement over the reconstruction-only backbone (31.78\,dB).
This confirms, on a second dataset, that the motion factorization acts as a beneficial inductive bias rather than a reconstruction penalty, consistent with the HyperNeRF finding in the main paper (28.52 vs.\ 27.59\,dB, ${+}0.93$\,dB, Table~4 of the main paper).
The Neu3D improvement is larger (${+}1.45$ vs.\ ${+}0.93$\,dB), likely because multi-view Neu3D scenes provide more geometric constraints for the per-object SE(3) transforms to exploit.

The mean rigid-share ratio ($\bar{\rho} = 0.879$) is substantially higher than on HyperNeRF ($\bar{\rho} = 0.472$), reflecting the fact that Neu3D scenes contain predominantly rigid object manipulations (cutting, searing, pouring) with less non-rigid deformation than the HyperNeRF scenes.

All architecture, hyperparameters, and training schedules are identical to the HyperNeRF experiments; no per-scene or per-dataset tuning was applied.

% ═════════════════════════════════════════════════════════
\section{Failure Cases}
\label{app:failures}

\noindent\secchip{colMotion}\enspace\textbf{F1: Systematic mask errors.}
Consistent mis-segmentation across all views propagates into incorrect transforms.

\noindent\secchip{colLang}\enspace\textbf{F2: Appearance-driven states.}
Kinematic features uninformative when state changes are visual, not motoric (chickchicken).

\noindent\secchip{colLang}\enspace\textbf{F3: Long occlusion.}
Embedding degrades to $\be_k^{\mathrm{static}}$.

\noindent\secchip{colMotion}\enspace\textbf{F4: Articulation.}
SE(3) captures dominant rigid component only; part-level motion absorbed into residuals.

\noindent\secchip{colSync}\enspace\textbf{F5: Scale ambiguity.}
Per-object standardization mitigates single-object features but not relational dims~20--27.

\smallskip
\noindent Foundation model biases (SAM~3, SigLIP) are inherited; relevant for safety-critical deployment.

% ═════════════════════════════════════════════════════════
\section{Broader Impact}
\label{app:broader_impact}

\noindent\secchip{colSync}\enspace
The central premise of this work is that reconstruction, object-factored motion, and language are not independent problems to be solved in sequence but coupled aspects of a single representational challenge.
By demonstrating that a lightweight in-loop factorization can expose interpretable motion primitives at negligible reconstruction cost, and that a closed-form linear map from those primitives to language embeddings more than doubles temporal localization quality over post-hoc distillation baselines, the method provides empirical evidence that \emph{motion structure is a viable and underexploited conditioning signal for semantic grounding in dynamic scenes}.

This finding points toward a broader research direction: building 4D representations in which every learned quantity (geometry, deformation, object identity, semantics) is structurally accessible to every other, rather than layered as frozen modules.
The object-time language field, the kinematic feature interface, and the per-object ridge map are intentionally simple instantiations of this principle; stronger motion models, richer feature spaces, and tighter integration with world models and embodied planners are natural next steps that the modular architecture is designed to support.

\end{document}